\documentclass[reviewcopy]{elsart}

\usepackage{graphicx}
\usepackage{tabularx}
\usepackage{colortbl}
\usepackage{setspace}

\usepackage{subfigure}
\usepackage{url}
\usepackage{amsmath}
\usepackage{amssymb}
\usepackage{algorithm}
\usepackage{algorithmic}
\usepackage{array}
\usepackage{longtable}
\usepackage{multirow}

\usepackage{wasysym}

\graphicspath{{imgs-eps/}}

\newcolumntype{V}{>{$\vcenter\bgroup\hbox\bgroup}c<{\egroup\egroup$}}

\begin{document}

\begin{frontmatter}

\title{\vspace{-30pt}\hspace{-20pt}~Infrared Face Recognition: A Comprehensive Review of Methodologies and Databases\vspace{-30pt}}





Reza~Shoja~Ghiass$^a$~~
Ognjen~Arandjelovi\'c$^b$~~
Abdelhakim~Bendada$^a$~~
Xavier~Maldague$^a$

{\small
\begin{tabular}{ll}
&\\[-25pt]
$^a$ Computer Vision \& Systems & $^b$ Pattern Recognition \& Data Analytics\\[-20pt]
\hspace{8pt}Universit\'{e} Laval, Canada  & \hspace{8pt}Deakin University, Australia\\[-20pt]
& \hspace{8pt}\texttt{ognjen.arandjelovic@gmail.com}\\[-20pt]
& \hspace{8pt}+61(0)3 93955628\\[-10pt]
\end{tabular}}

\title{}
\author{}

\renewcommand{\baselinestretch}{1.76}

\vspace{-145pt}
\begin{abstract}
\vspace{-35pt}\\
\hspace{-12pt}Automatic face recognition is an area with immense practical potential which includes a wide range of commercial and law enforcement applications. Hence it is unsurprising that it continues to be one of the most active research areas of computer vision. Even after over three decades of intense research, the state-of-the-art in face recognition continues to  improve, benefitting from advances in a range of different research fields such as image processing, pattern recognition, computer graphics, and physiology. Systems based on visible spectrum images, the most researched face recognition modality, have reached a significant level of maturity with some practical success. However, they continue to face challenges in the presence of illumination, pose and expression changes, as well as facial disguises, all of which can significantly decrease recognition accuracy. Amongst various approaches which have been proposed in an attempt to overcome these limitations, the use of infrared (IR) imaging has emerged as a particularly promising research direction. This paper presents a comprehensive and timely review of the literature on this subject. Our key contributions are: (i) a summary of the inherent properties of infrared imaging which makes this modality promising in the context of face recognition, (ii) a systematic review of the most influential approaches, with a focus on emerging common trends as well as key differences between alternative methodologies, (iii) a description of the main databases of infrared facial images available to the researcher, and lastly (iv) a discussion of the most promising avenues for future research.
\end{abstract}
~
\end{frontmatter}

\setstretch{1.90}

\section{Introduction}\label{s:intro}
In the last two decades automatic face recognition has consistently been one of the most active research areas of computer vision and applied pattern recognition. Systems based on images acquired in the visible spectrum have reached a significant level of maturity with some practical success \cite{KongHeoAbidPaik+2005}. However, a range of nuisance factors continue to pose serious problems when visible spectrum based face recognition methods are applied in a real-world setting. Dealing with illumination, pose and facial expression changes, and facial disguises is still a major challenge.

There is a large corpus of published work which has attempted to overcome the aforesaid difficulties by developing increasingly sophisticated models which were then applied on the same type of data -- usually images acquired in the visible spectrum (wavelength approximately pin the range $390-750$~nm). Pose, for example, has been normalized by a learnt 2D warp of an input image \cite{GrosMattBake2006}, generated from a model fitted using an analysis-by-synthesis approach \cite{BlanVett2003} or synthesized using a statistical method \cite{MohaPrinKaut2009}, while illumination has been corrected for using image processing filters \cite{NishYama2006} and statistical facial models \cite{WolfShas2003}, amongst others, with varying levels of success. Other methods adopt a multi-image approach by matching sets \cite{ChinSute2006,LuiBeve2008,FanYeun2006,AranCipo2004a} or sequences of images \cite{LeeKrie2005,BowyChanFlynChen2006}. Another increasingly active research direction has pursued the use of alternative modalities. For example, it is clear that data acquired using 3D scanners \cite{PanWu2005,GodiRessGrot2004} is inherently robust to illumination and pose changes. However, the cost of these systems is high and the process of data collection overly restrictive for most practical applications.

\subsection{Infrared Spectrum}
Infrared imagery is a modality which has attracted particular attention, in large part due to its invariance to the changes in illumination by visible light \cite{ZouKittMess2007}. A detailed account of the relevant physics, which is outside the scope of this paper, can be found in \cite{Mald2001}. In the context of face recognition, data acquired using infrared cameras has distinct advantages over the more common cameras which operate in the visible spectrum. For instance, infrared images of the faces can be obtained under any lighting condition, even in completely dark environments, and there is some evidence that thermal infrared (see Sec.~\ref{ss:composition}) ``appearance'' may exhibit a higher degree of robustness to facial expression changes~\cite{FrieYesh2003}. Thermal infrared energy is also less affected by scattering and absorption by smoke or dust than reflected visible light~\cite{ChanKoscAbidKong+2008a,NicoSchm2011}. Unlike visible spectrum imaging, infrared imaging can be used to extract not only exterior, but also useful subcutaneous anatomical information, such as the vascular network of a face~\cite{BuddPavlTsia2005}. Finally, in contrast to visible spectrum imaging, thermal vision can be used to detect facial disguises \cite{PavlSymo2000}.

\subsection{Spectral Composition}\label{ss:composition}
\begin{figure}[htb]
  \label{f:bodyheat}
  \centering
  \includegraphics[width=0.5\textwidth]{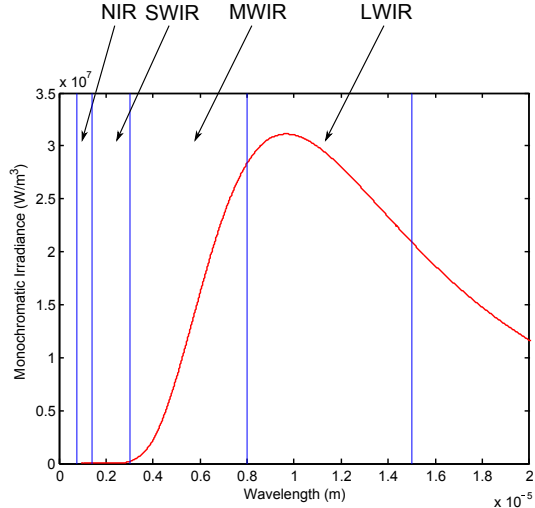}
   \caption{ The idealized spectrum of heat emission by the human body predicted by Planck's law at 305~K, with marked boundaries of the four infrared sub-bands of interest in this paper: near-wave (NIR), short-wave (SWIR), medium-wave (MWIR) and long-wave (LWIR). Observe that the emission in the NIR and SWIR sub-bands is nearly zero. As a consequence, imaging in these bands is by necessity active i.e.\ it requires an illuminator at the appropriate wavelengths. }
\end{figure}

In the existing literature, it has been customary to divide the infrared spectrum into four sub-bands: near IR (NIR; wavelength $0.75-1.4\mu$m), short wave IR (SWIR; wavelength $1.4-3\mu$m), medium wave IR (MWIR; wavelength $3-8\mu$m), and long wave IR (LWIR; wavelength $8-15\mu$m). This division of the IR spectrum is also observed in the manufacturing of infrared cameras, which are often made with sensors that respond to electromagnetic radiation constrained to a particular sub-band. It should be emphasized that the division of the IR spectrum is not arbitrary. Rather, different sub-bands correspond to continuous frequency chunks of the solar spectrum which are divided by absorption lines of different atmospheric gasses~\cite{Mald2001}. In the context of face recognition, one of the largest differences between different IR sub-bands emerges as a consequence of the human body's heat emission spectrum which is, in its idealized form, shown in Fig.~\ref{f:bodyheat}. Specifically, note that most of the heat energy is emitted in LWIR sub-band, which is why it is often referred to as the thermal sub-band (this term is sometimes extended to include the MWIR sub-band). Significant heat is also emitted in the MWIR sub-band. Both of these sub-bands can be used to \emph{passively} sense facial thermal emissions without an external source of light. This is one of the reasons why LWIR and MWIR sub-bands have received the most attention in the face recognition literature. In contrast to them, facial heat emission in the SWIR and NIR sub-bands is small and recognition systems operating on data acquired in these sub-bands require appropriate illuminators (invisible to the human eye) i.e.\ recognition is \emph{active} in nature~\cite{ZouKittMess2005}. In recent years, the use of NIR also started received increasing attention from the face recognition community, while the utility of the SWIR sub-band has yet to be studied in depth.


\subsection{Challenges}
\begin{figure}[thp]
  \centering
  \subfigure[]{\includegraphics[width=0.25\textwidth]{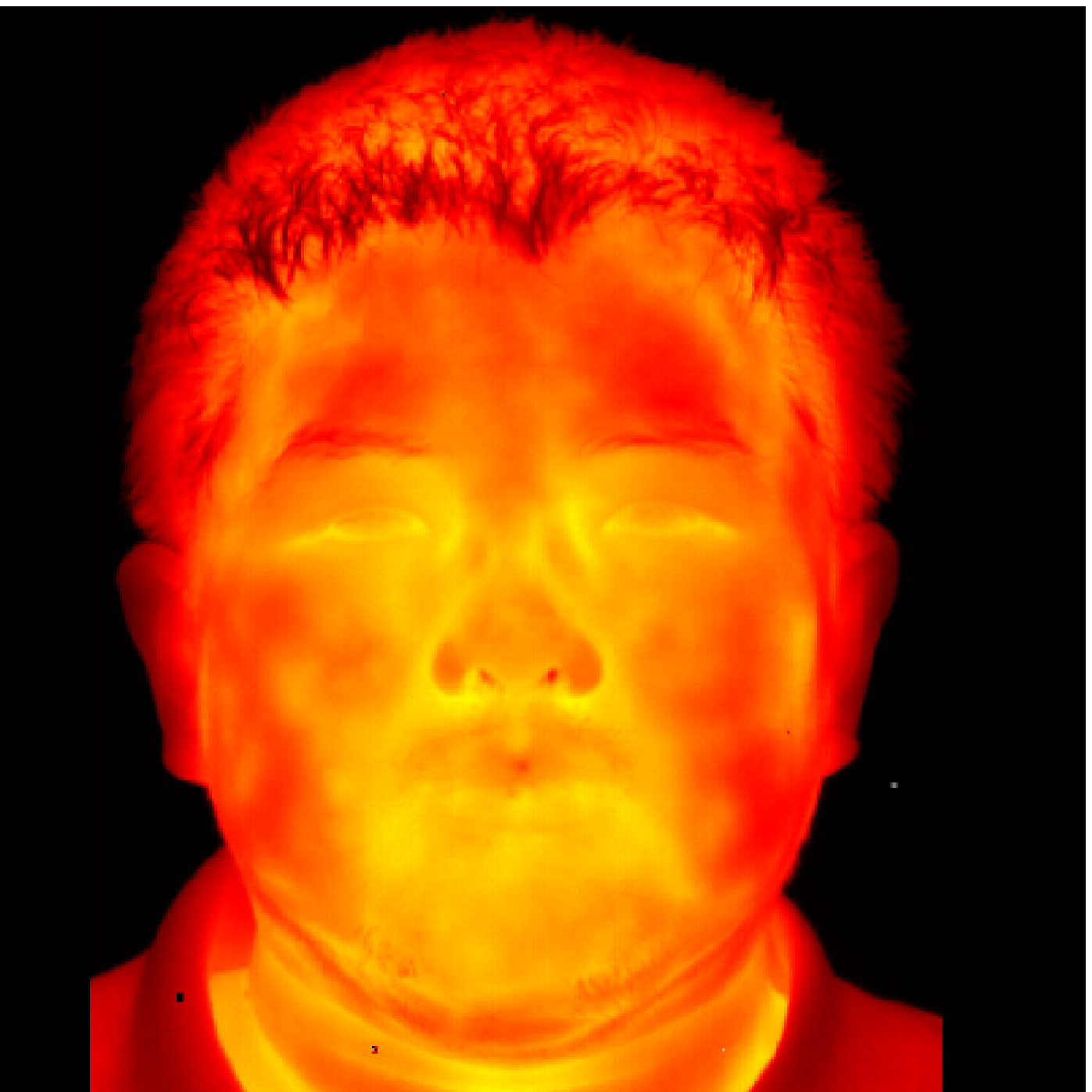}}~~~~~
  \subfigure[]{\includegraphics[width=0.25\textwidth]{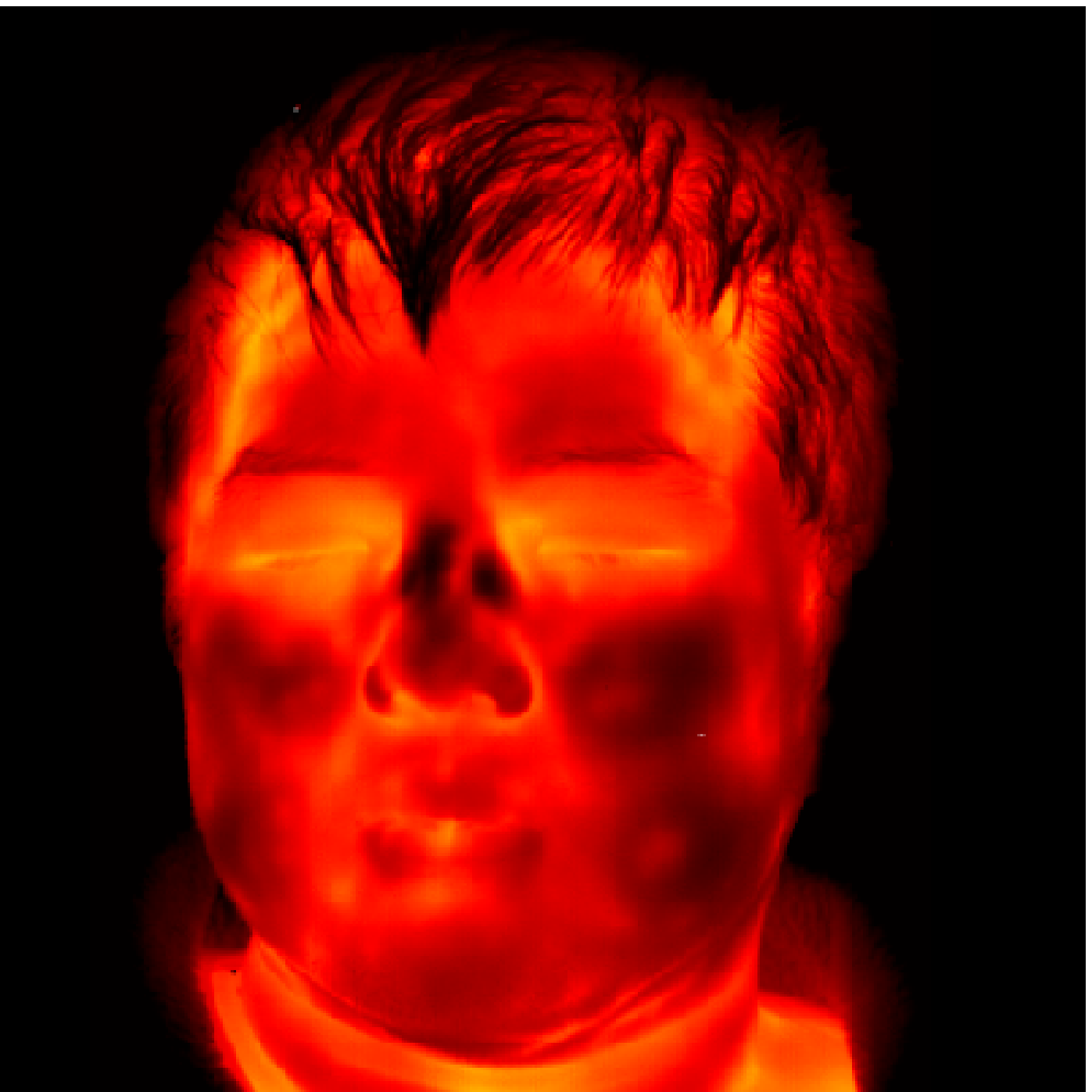}}
  \caption{ Thermal IR images of a person acquired during the course of an average day (a), and following exposure to cold (b). Note that the images were enhanced and are shown in false colour for easier visualization. }
  \label{f:phys}
\end{figure}

The use of infrared images for automatic face recognition is not void of challenges. For example, MWIR and LWIR images are sensitive to the environmental temperature, as well as the emotional, physical and health condition of the subject, as illustrated in Fig.~\ref{f:phys}. They are also affected by alcohol intake. Another potential problem is that eyeglasses are opaque to the greater part of the IR spectrum (LWIR, MWIR and SWIR) \cite{TasmJaeg2009}. This means that a large portion of the face wearing eyeglasses may be occluded, causing the loss of important discriminative information. Unsurprisingly, each of the aforementioned challenges has led to and motivated a new research direction. Some researchers have suggested fusing the information from IR and visible modalities as a possible solution to the problem posed by the opaqueness of eyeglasses \cite{KongHeoAbidPaik+2005}. Others have described methods which use thermal infrared images to extract a range of invariant features such as facial vascular networks \cite{BuddPavlTsia2005,BuddPavlTsiaBaza2007} or blood perfusion data \cite{WuWeiFangLi+2007} in order to overcome the temperature dependency of thermal ``appearance''. Another consideration of interest pertains to the impact of sunlight if recognition is performed outdoors and during daytime.  Although invariant to the changes in the illumination by visible light itself (by definition), the infrared ``appearance'' in the NIR and SWIR sub-bands \emph{is} affected by sunlight which has significant spectral components at the corresponding wavelengths. This is one of the key reasons why NIR and SWIR based systems which perform well indoors struggle when applied outdoors \cite{LiChuLiaoZhan2007,WuSongJianXie+2005}.

\subsection{Aims and Organization}
The aim of this paper is to present a thorough literature review of the growing and increasingly important problem of infrared face recognition. In comparison with the already published reviews of the field, by Kong \textit{et al.}\ \cite{KongHeoAbidPaik+2005}, Akhloufi \textit{et al.}\ \cite{AkhlBendBats2008} and Ghiass \textit{et al.}\ \cite{GhiaBendMald2010,GhiaAranBendMald2013b}, the present paper makes several important contributions. Firstly, we survey a much greater corpus of relevant work. What is more, we include and give particular emphasis to the most recent advances in the field. As such, our review is both the most comprehensive and the most up-to-date review of infrared based face recognition to date. Finally, our work is distinguished from other reviews of the field also by its original categorization of different methodologies, which adds further insight into the evolution of dominant research trends.

The remainder of this paper is organized as follows. Firstly, the inherent advantages and disadvantages of infrared data in the context of face recognition are discussed in Sec.~\ref{s:ir}. Sec.~\ref{s:main} comprises the main part of the paper. This is where we describe different recognition approaches proposed in the literature, grouped by the methodology or the type of features employed for recognition. Sec.~\ref{s:dbs} which follows aims to survey various databases of infrared facial images. Our focus was on free databases, but a number of proprietary databases which have gained prominence through important peer-reviewed publications are included as well. Finally, the most important conclusions and trends in the field to date are summarized in Sec.~\ref{s:summary}.

\section{Infrared Data: Advantages and Disadvantages}\label{s:ir}
Many of the methods for infrared based face recognition have been inspired by or are verbatim copies of algorithms which were initially developed for visible spectrum recognition. In most cases, these methods make little use of the information about the spectrum which was used to acquire images. However, the increasing appreciation of challenges encountered in trying to robustly match infrared images strongly suggests that domain specific properties of data should be exploited more. Indeed, as we discuss in Sec.~\ref{s:main} and~\ref{s:summary}, the recent trend in the field has been moving in this direction, increasingly complex IR specific models being proposed. Thus, in this section we focus on the relevant differences of practical significance between infrared and visible spectrum images. The use of infrared imagery provides several important advantages as well as disadvantages, and we start with a summary of the former first.


\subsection{Advantages of Infrared Data in Automatic Face Recognition}
Much of the early work on the potential of infrared images as identity signatures was performed by Prokoski \textit{et al.}\ \cite{ProkRiedCoff1992,Prok1992a,ProkRied1998}. They were the first to advance the idea that infrared ``appearance'' could be used to extract robust biometric features which exhibit a high degree of uniqueness and repeatability.

\begin{figure}[htb]
  \centering
  \subfigure[]{\includegraphics[width=0.45\textwidth,height=0.1125\textwidth]{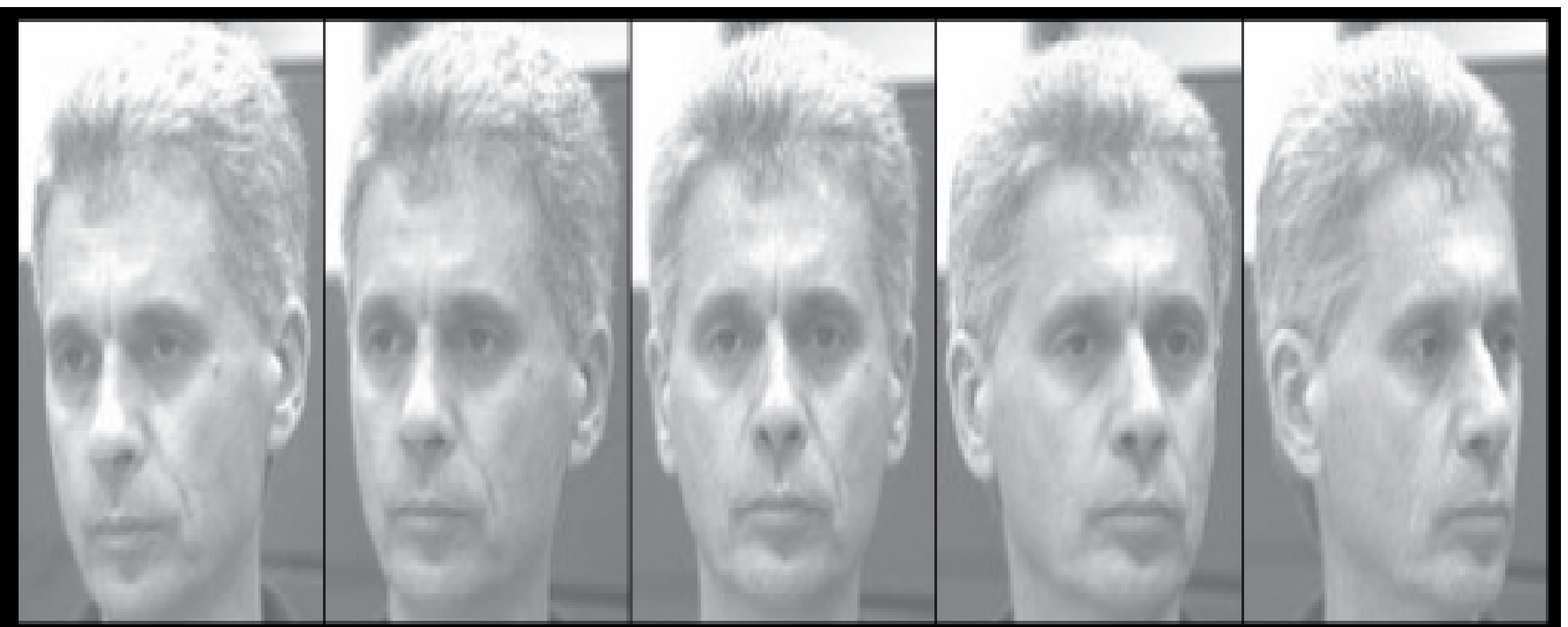}}~~~~~
  \subfigure[]{\includegraphics[width=0.45\textwidth,height=0.1125\textwidth]{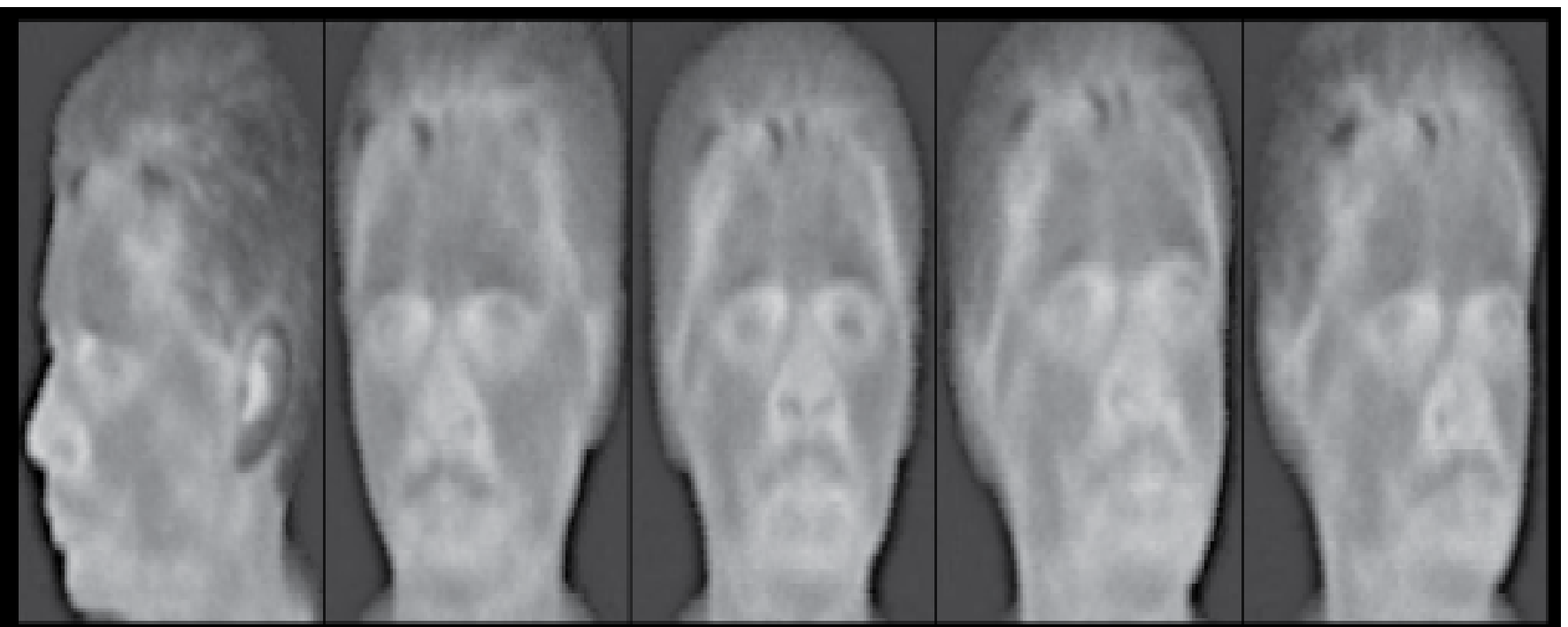}}
  \caption{ Examples of (a) visible spectrum images and (b) the corresponding thermograms of an individual across different poses/views \cite{FrieYesh2003}. Note that the visible and thermal images were not acquired concurrently so the poses in (a) and (b) are not exactly the same.  }
  \label{f:visVsIR}
\end{figure}

Facial expression and pose changes are two key factors that a face recognition system should be robust to for it to be useful in most practical applications of interest. By comparing image space differences of thermal and visible spectrum images, Friedrich \textit{et al.}\ \cite{FrieYesh2003} found that thermal images are less affected by changes in pose or facial expression than their visible spectrum counterparts. An example is shown in Fig.~\ref{f:visVsIR}. Illumination invariance of different infrared sub-bands was analyzed in detail by Wolff \textit{et al.}\ \cite{WolfSocoEvel2001} who showed the superiority of infrared over visible data with respect to this important nuisance variable.

The very nature of thermal imaging also opens the possibility of non-invasive extraction and use of superficial anatomical information for recognition. Blood vessel patterns are one such example. As they continually transport circulating blood, blood vessels are somewhat warmer than the surrounding tissues. Since thermal cameras capture the heat emitted by a face, standard image processing techniques can be readily used to extract blood vessel patterns from facial thermograms. An important property of these patterns which makes them particularly attractive for use in recognition is that the blood vessels are ``hardwired'' at birth and form a pattern which remains virtually unaffected by factors such as aging, except for predictable growth \cite{PersBusc2011}. Moreover, it appears that the human vessel pattern is robust enough to facilitate scaling up to large populations \cite{ProkRied1998}. Prokoski \textit{et al.}\ estimate that about 175 blood vessel based minutiae can be extracted from a full facial image \cite{ProkRied1998} which, they argued, can exhibit a far greater number of possible configurations than the size of the foreseeable maximum human population. It should be noted that the authors did not propose a specific algorithm to extract the minutiae in question. In the same work, the authors also argued that forgery attempts and disguises can both be detected by infrared imaging. The key observation is that the temperature distribution of artificial facial hair or other facial wear differs from that of natural hair and skin, allowing them to be differentiated one from another.

\subsubsection{The Twin Paradox}
An interesting question first raised by Prokoski \textit{et al.}\ \cite{ProkRied1998} concerns thermograms of monozygotic twins. The appearance of monozygotic twins (or ``identical'' in common vernacular) is nearly identical in the visible spectrum. Using a small number of thermograms of monozygotic twins which were qualitatively assessed for similarity, Prokoski \textit{et al.}\ found that the difference in appearance was significantly greater in the thermal than in the visible spectrum, and sufficiently so to allow for them to be automatically differentiated. This hypothesis was disputed by subsequent contradictory findings of Chen \textit{et al.}\ \cite{ChenFlynBowy2005}. However, the weight of evidence provided both by Prokoski \textit{et al.}\ as well as Chen \textit{et al.}\ is inadequate to allow for a confident conclusion to be made. Both positive and negative claims are based on experiments which use little data and lack sufficient rigour. In addition, it is plausible that the truth may be somewhere in the middle, that is, that in some cases monozygotic twins can be differentiated from their thermograms and in others not, depending on a host of physiological variables.

\subsection{Limitations of Infrared Data in Automatic Face Recognition}
\begin{figure}[htb]
  \centering
  \footnotesize
  \begin{tabular}{VVV}
    \begin{tabular}{VV}
     \includegraphics[width=0.18\textwidth,height=0.12\textwidth]{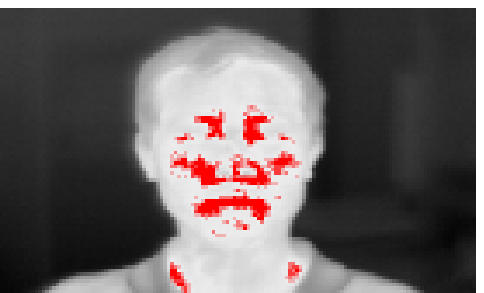}&
     \includegraphics[width=0.18\textwidth,height=0.12\textwidth]{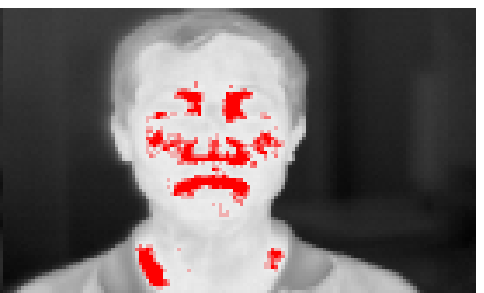}\\
     &\\[-15pt]
     (a) 28.4$^{\circ}$C & (b) 28.7$^{\circ}$C\\
     &\\[-15pt]
     \includegraphics[width=0.18\textwidth,height=0.12\textwidth]{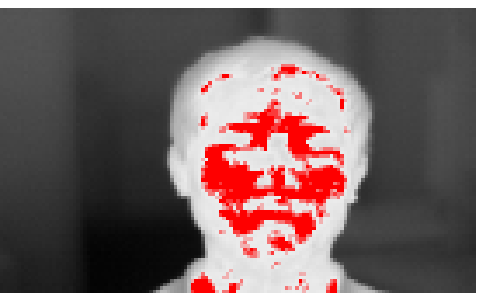}&
     \includegraphics[width=0.18\textwidth,height=0.12\textwidth]{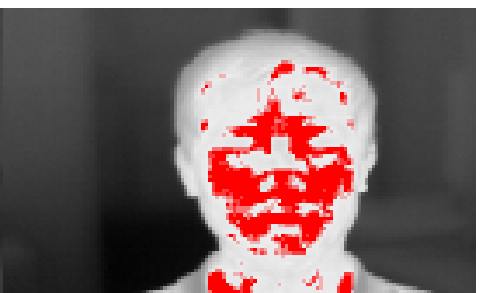}\\
     &\\[-15pt]
     (c) 28.9$^{\circ}$C & (d) 29.3$^{\circ}$C\\
    \end{tabular}&~\hspace{25pt}~&
    \begin{tabular}{VV}
     \includegraphics[width=0.18\textwidth]{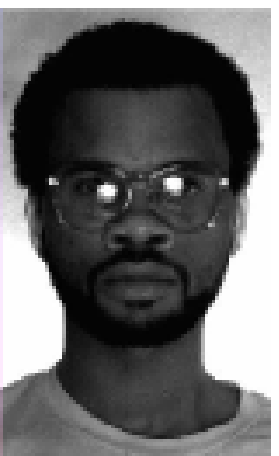} & \includegraphics[width=0.18\textwidth]{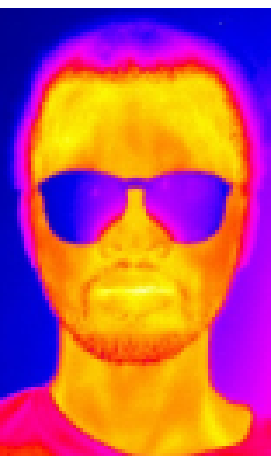}\\
     &\\[-5pt]
     (e) Visible & (f) Thermal\\
    \end{tabular}
  \end{tabular}
  \caption{ (a--d) Thermal infrared images of the same person taken at different ambient temperatures \cite{WuFangXieLian2008}. Regions marked in red correspond to heat intensity values exceeding 93\% of the maximal heat value representable in the images. (e,f) A corresponding pair of visual and false colour thermal images of a person wearing eyeglasses. Notice the complete loss of information around the eyes in the thermal image. The visible spectrum image is affected much less: some information is lost due to localized specular effects and the occlusion of the face by the frame of the eyeglasses. }
  \label{f:ambient}
\end{figure}

In the context of automatic face recognition, the main drawback specific to the thermal sub-band images (or \emph{thermograms}, as they are often referred to), the most often used sub-band of the infrared spectrum, stems from the fact that the heat pattern emitted by the face is affected by a number of confounding variables, such as ambient temperature, air flow conditions, exercise, postprandial metabolism, illness and drugs \cite{ProkRied1998}. Sensitivity to ambient temperature is illustrated on an example in Fig.~\ref{f:ambient}~(a--d). Some of the confounding variables produce global, others local thermal appearance changes. Wearing clothes, experiencing stress, blushing, having a headache or an infected tooth are examples of factors which can effect localized changes.

The high sensitivity of the facial thermogram to a large number of extrinsic factors makes the task of finding persistent and discriminative features a challenging one. It also lends support to the ideas first voiced by Prokoski \textit{et al.}\ who argued against the use of thermal appearance based methods in favour of anatomical feature based approaches invariant to many of the aforementioned factors. As we will discuss in Sec.~\ref{ss:features}, this direction of infrared based face recognition has indeed attracted a substantial research effort.

Another drawback of using the infrared spectrum for face recognition is that glass and thus eyeglasses are opaque to wavelengths longer and including the SWIR sub-band. Consequently an important part of the face, one rich in discriminative information, may be occluded in the corresponding images. In particular, the absence of appearance information around the eyes can greatly decrease recognition accuracy~\cite{AranHammCipo2010}. Multi-modal fusion based methods have been particularly successful in dealing with this problem, as described in detail in Sec.~\ref{sss:glasses}.

Lastly, a major challenge when NIR and SWIR sub-bands are used for recognition, stems from their sensitivity to sunlight which has significant spectral components at the corresponding wavelengths \cite{LiChuLiaoZhan2007,WuSongJianXie+2005}. In this sense, the problem of matching images acquired in NIR and SWIR sub-bands is similar to matching visible spectrum images.

\section{Face Recognition Using Infrared}\label{s:main}
In this review, we recognize four main groups of face recognition methodologies which use infrared data: holistic appearance based, feature based, multi-spectral based, and multi-modal fusion based. Holistic appearance methods use the entire infrared appearance image of a face for recognition. Feature based approaches use infrared images to extract salient face features, such as facial geometry, its vascular network or blood perfusion data. Spectral model based approaches model the process of infrared image formation to decompose images of faces. Some approaches directly use data from multi-spectral or hyper-spectral imaging sensors to obtain facial images across different frequency sub-bands. Multi-modal fusion based approaches combine information contained in infrared images with information contained in other types of modalities, such as visible spectrum data, with the aim of exploiting their complementary advantages. As the understanding of the challenges of using infrared data for face recognition has increased, this direction of research has become increasingly active.

\subsection{Appearance-Based Methods}\label{ss:appearance}
\begin{figure}[htb]
  \centering
  \includegraphics[width=0.2\textwidth]{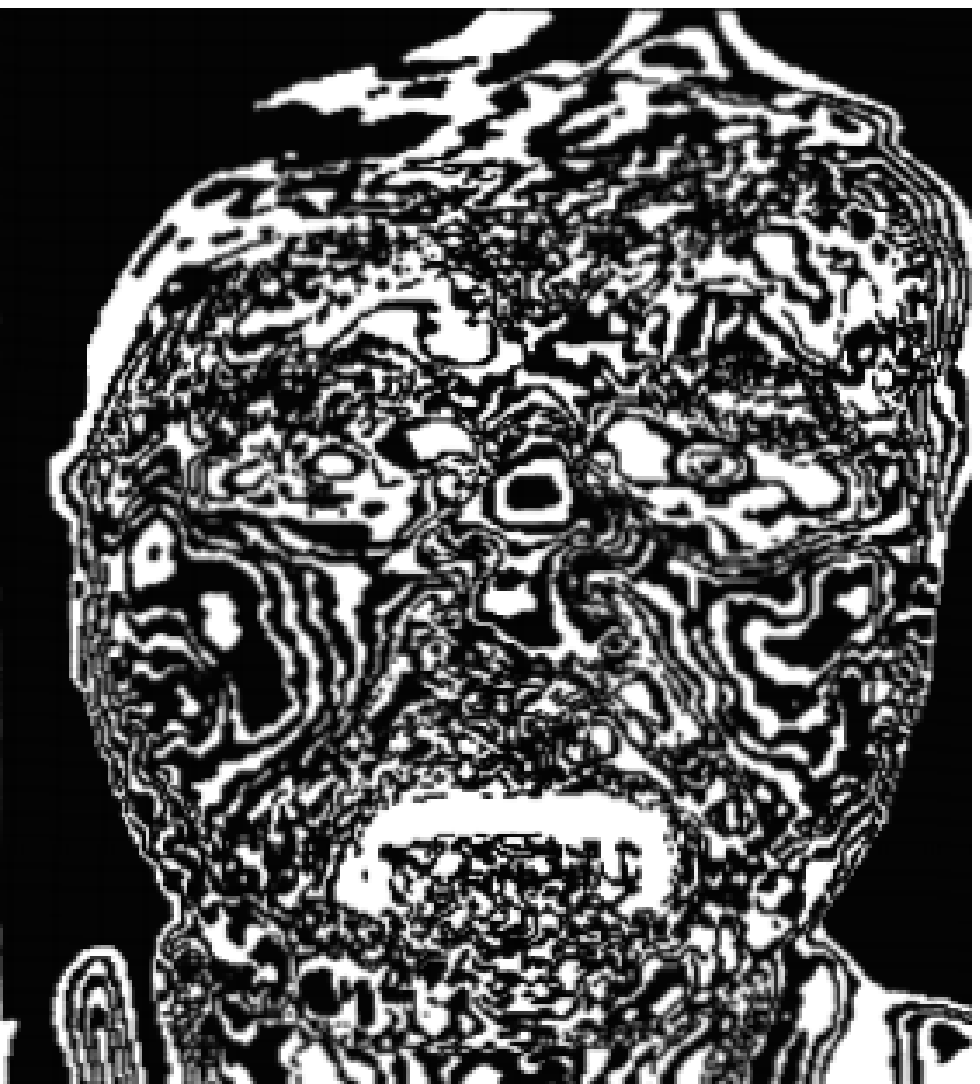}\hspace{15pt}
  \includegraphics[width=0.2\textwidth]{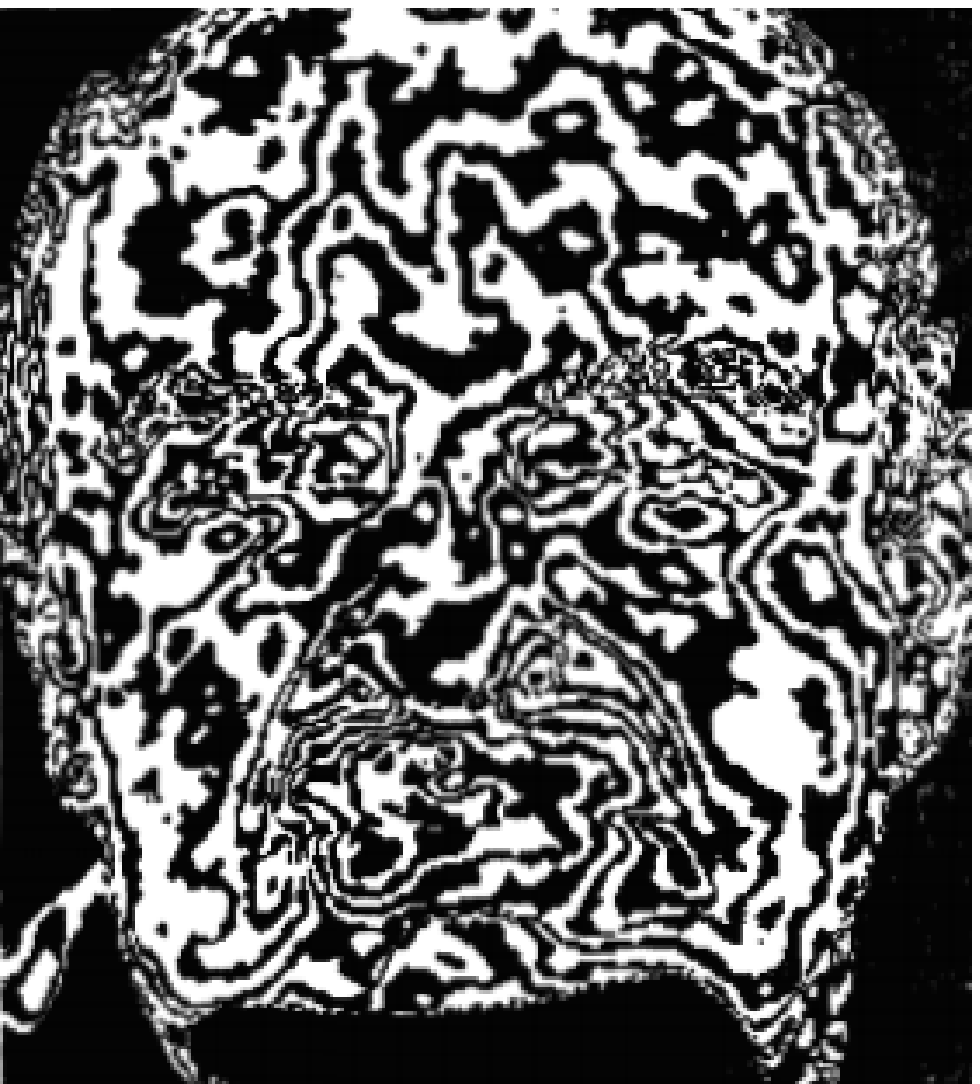}\hspace{15pt}
  \includegraphics[width=0.2\textwidth]{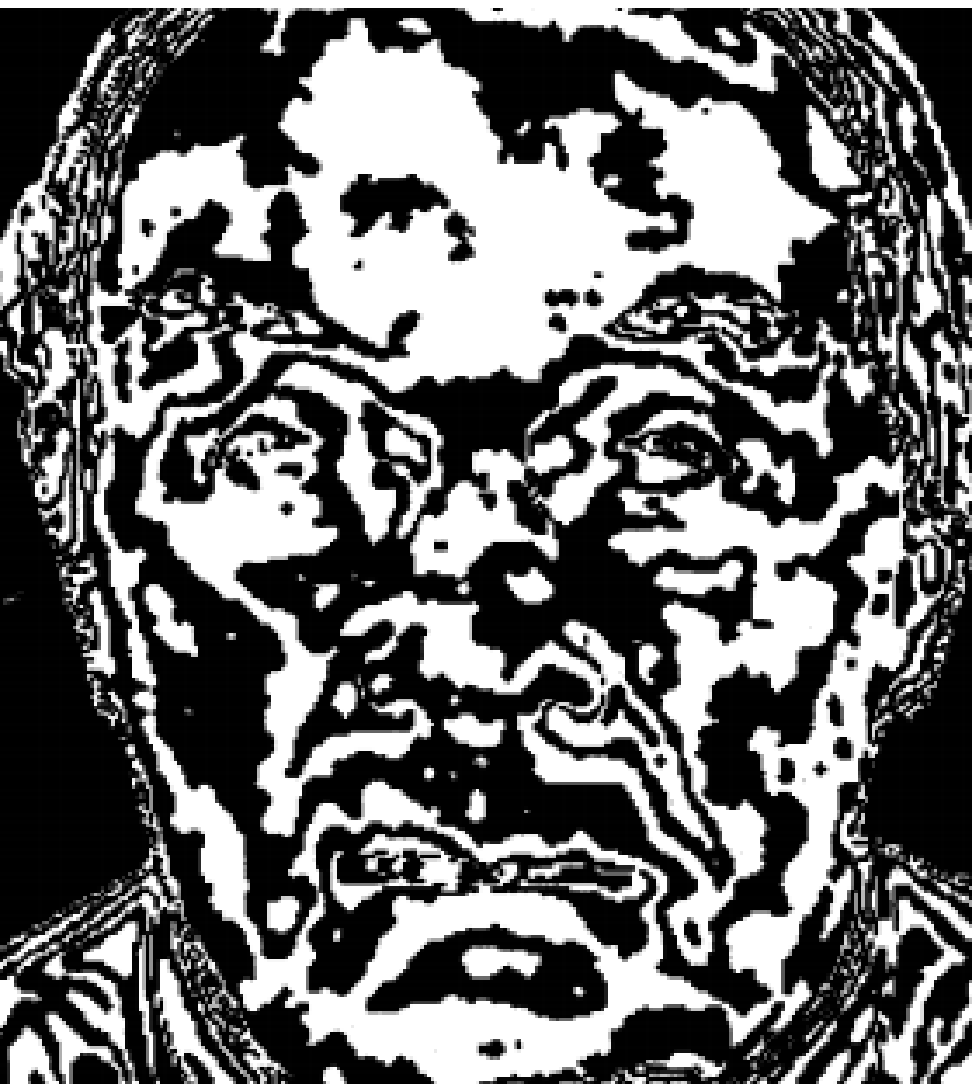}\hspace{15pt}
  \includegraphics[width=0.2\textwidth]{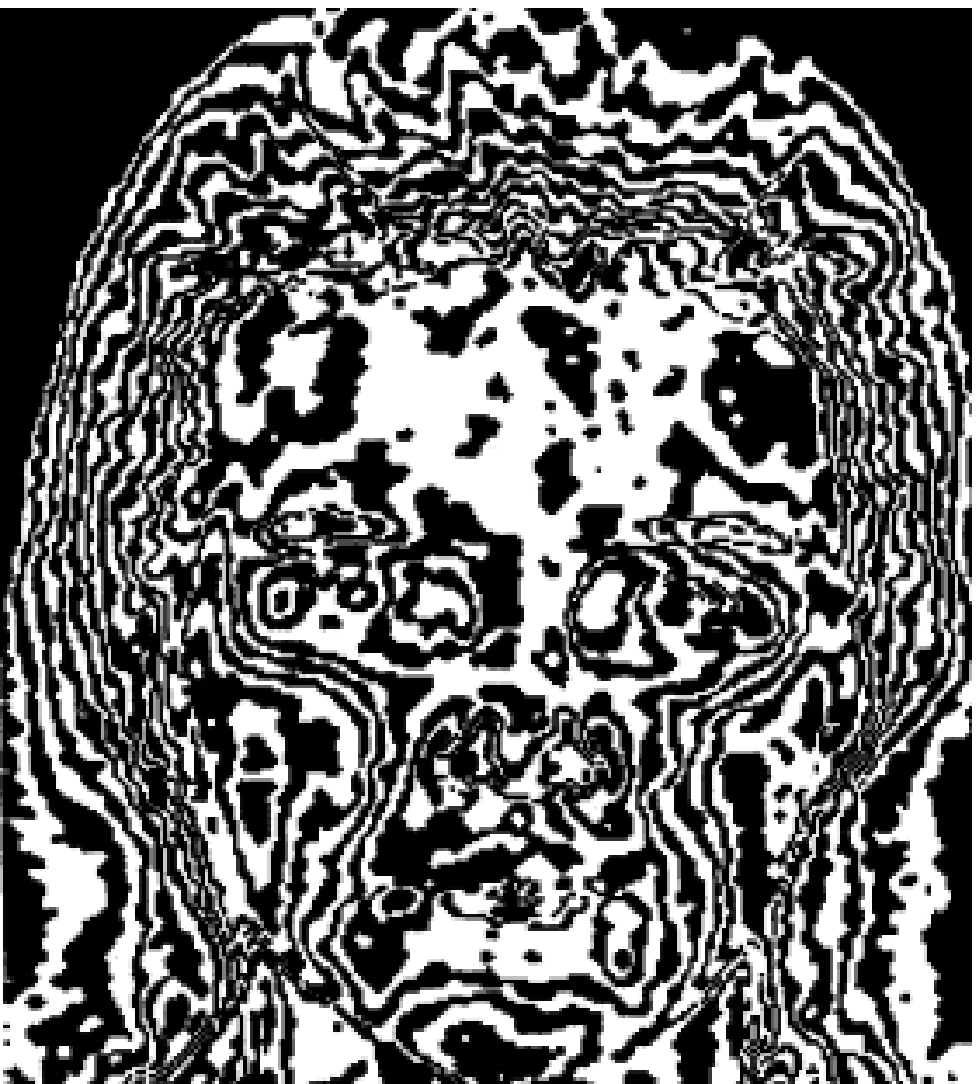}\hspace{15pt}
  \caption{ Images of ``elementary shapes'' proposed by Prokoski \textit{et al.}\ \cite{ProkRiedCoff1992}. }
  \label{f:prokoski}
\end{figure}

The earliest attempts at examining the potential of infrared imaging for face recognition dates back to 1992 and the work done by Prokoski \textit{et al.}\ \cite{ProkRiedCoff1992}. Their work introduced the concept of ``elementary shapes'' extracted from thermograms, which are likened to fingerprints. While precise technical detail of the method used to extract these elementary shapes is lacking, it appears that they are isothermal regions segmented out from an image, as illustrated in Fig.~\ref{f:prokoski}. There is no published record on the effectiveness of this representation.

\subsubsection{Early Approaches}\label{ss:early}
Perhaps unsurprisingly, most of the automatic methods which followed the work of  Prokoski \textit{et al.}\ closely mirrored in their approach methods developed for the more popular visible spectrum based recognition. Generally, these used holistic face appearance in a simple statistical manner, with little attempt to achieve any generalization, relying instead on the availability of training data with sufficient variability of possible appearance for each subject.

One of the first attempts at using infrared data in an automatic face recognition system was described by Cutler \cite{Cutl1996}. His method was entirely based on the popular Eigenfaces method proposed by Turk and Pentland \cite{TurkPent1991}. Using a database of 288 thermal images (12 images for each of the 24 subjects in the database) which included limited pose and facial expression variation, Cutler reported rank-1 recognition rates of 96\% for frontal and semi-profile views, and 100\% for profile views. These recognition rates compared favourably with those achievable using the same methodology on visible spectrum images. Following these promising results, many of the subsequently developed algorithms also adopted Eigenfaces as the baseline classifier. For example, findings similar to those made by Cutler were independently reported by Socolinsky \textit{et al.}\ \cite{SocoWolfNeuhEvel2001}.

In their later work, Socolinsky \textit{et al.}\ \cite{SocoSeli2004} and Selinger \textit{et al.}\ \cite{SeliSoco2002,SeliSoco2004} extended their comparative evaluation of thermal and visible data based recognition using a wider range of linear methods: Eigenfaces (that is, principal component analysis), linear discriminant analysis, local feature analysis and independent component analysis. Their results corroborated previous observations made in the literature on the superiority of the thermal spectrum for recognition in the presence of a range of nuisance variables. However, the conclusions that could be drawn from their analysis of different recognition approaches, or indeed that of Culter, were limited by the insufficiently challenging data sets which were used: pose and expression variability was small, training and test data were acquired in a single session, and the subjects wore no eyeglasses. This is reflected in the fact that all of the evaluated algorithms achieved comparable, and in practical terms high, recognition rates (approximately 93-98\%).

\subsubsection{Effects of Registration}\label{ss:registration}
In practice, after detection faces are still insufficiently well aligned (registered) for pixel-wise comparison to be meaningful. The simplest and the most direct way of registering faces is by detecting a discrete set of salient facial features and then applying a geometric warp to map them into a canonical frame. Unlike in the case of images acquired in the visible spectrum, in which several salient facial features (such as the eyes and the mouth) can usually be reliably detected \cite{ChriCootScot2004,AranZiss2005,DingMart2010}, most of the work to date supports the conclusion that salient facial feature localization in thermal images is significantly more challenging. Different approaches, which mainly focus on the eyes, were described by Tzeng \textit{et al.}\ \cite{TzenLeeChen2011}, Arandjelovi\'c \textit{et al.}\ \cite{AranHammCipo2010}, Jin \textit{et al.}\ \cite{JinShouXiuGu2009}, Bourlai \textit{et al.}\ \cite{BourJafr2011,BourWhitKaka2011} and Martinez \textit{et al.}\ \cite{MartBinePant2010}. What is more, the effect of feature localization errors and thus registration errors seems to be greater for thermal than visible spectrum images. This was investigated by Chen \textit{et al.}\ \cite{ChenFlynBowy2003} who demonstrated a substantial reduction in thermal based recognition rates when small localization errors were synthetically introduced to manually marked eye positions.

Zhao \textit{et al.}\ \cite{ZhaoGrig2005} circumvent the problem of localizing the eyes in passively acquired images by their use of additional active NIR data. A NIR lighting source placed close to and aligned with the camera axis is used to illuminate the face. Because the interior of the eyes reflects the incident light the pupils appear distinctively bright and as such are readily detected in the observed image (the so-called ``bright pupil'' effect). Zhao \textit{et al.}\ use the locations of pupils to register images of faces, which are then represented using their DCT coefficients and classified using a support vector machine. A related approach has also been described by Zou \textit{et al.}\ \cite{ZouKittMess2006}.

\subsubsection{Recent Advances in IR Appearance Based Recognition}\label{ss:modern}
Although the general trend in the field has been way from appearance based approaches and in the direction of feature and model based methods, the former have continued to attract some research interest. Much like the initial work, the recent advances in appearance based IR face recognition has closely mirrored research in visible spectrum based recognition. Progress in comparison with the early work is mainly to be found in the use of more sophisticated statistical techniques. For example, Elguebaly and Bouguila \cite{ElguBoug2011} recently described a method based on a generalized Gaussian mixture model, the parameters of which are learnt from a training image set using a Bayesian approach. Although substantially more complex, this approach did not demonstrate a statistically significant improvement in recognition on the IRIS Thermal/Visible database (see Sec.~\ref{ss:iris}), both methods achieving rank-1 rate of approximately 95\%. Lin \textit{et al.}  \cite{LinWenrLiZhij2011} were the first to investigate the potential of the increasingly popular compressive sensing in the context of IR face recognition. Using a proprietary database of 50 persons with 10 images each person, their results provided some preliminary evidence for the superiority of this approach over wavelet based decomposition (also see Sec.~\ref{sss:wavelet}).

Considering that the development of appearance based methods has nearly exclusively focused on the use of more sophisticated statistical techniques (rather than the incorporation of data specific knowledge, say), it is a major flaw in this body of research that the data sets used for evaluation have not included the types of intra-personal variations that appearance based methods are likely to be sensitive to. Indeed, none of the data sets that we are aware of included intra-personal variations due to differing emotional states, alcohol intake or exercise, for example, or even ambient temperature. This observation casts a shadow on the reported results and impedes further development of algorithms which could cope with such variations in a realistic, practical setup.

\subsection{Feature-Based Methods}\label{ss:features}
An early approach which uses features extracted from thermal images, rather than raw thermal appearance, was proposed by Yoshitomi \textit{et al.}\ \cite{YoshMiyaTomiKimu1997}. Following the localization of a face in an image, their method was based on combining the results of neural network based classification of grey level histograms and locally averaged appearance, and supervised classification of a facial geometry based descriptor. The proposed method was evaluated across room temperature variations ranging from 302K to 285K. As expected, the highest recognition rates were attained (92\%+) when both training and test data were acquired at the same room temperature. However, the significant drop to 60\% for the highest temperature difference of 17K between training and test data demonstrated the lack of robustness of the proposed features and highlighted the need for the development of discriminative features exhibiting a higher degree of invariance to confounding variables expected in practice. Yoshitomi \textit{et al.}\ did not investigate the effectiveness of their method in the presence of other nuisance factors, such as pose or expression.

\subsubsection{Infrared Local Binary Patterns}\label{sss:lbps}
In a series of influential works, Li \textit{et al.}\ \cite{Li2005,LiChuAoZhan+2006,LiZhanLiaoZhu+2006a,LiChuLiaoZhan2007} were the first to use features based on local binary patterns (LBP) \cite{OjalPietHarw1994} extracted from infrared images. They apply their algorithm in an active setting which uses strong NIR light-emitting diodes, coaxial with the direction of the camera. This setup ensures both that the face is illuminated as homogeneously as possible, thus removing the need of algorithmic robustness to NIR illumination, as well as that the eyes can be reliably detected using the bright pupil effect. Evaluated in an indoor setting and with cooperative users, their system achieved impressive accuracy. However, as noted by Li \textit{et al.}\ \cite{LiChuLiaoZhan2007} themselves, it is unsuitable for uncooperative user applications or outdoor use due to the strong NIR component of sunlight (see Sec.~\ref{s:ir}).

The use of local binary patters was also investigated by Maeng \textit{et al.}\ \cite{MaenChoiParkLee+2011}, who applied them in a multi-scale framework on NIR imagery acquired at distance (up to 60m) with limited success, dense SIFT based features proving more successful in their recognition scenario. A good comparative evaluation of local binary patters in the context of a variety of linear and kernel methods was recently published by Goswami \textit{et al.}\ \cite{GoswChanWindKitt2011}.

\subsubsection{Wavelet Transform}\label{sss:wavelet}
Owing to its ability to capture both frequency and spatial information, the wavelet transform has been studied extensively as a means of representing a wide range of 1D and 2D signals, including face appearance in the visual spectrum. Srivastava \textit{et al.}\ \cite{SrivLiuThomHesh2001,SrivLiu2003} were the first to investigate the use of wavelets for extracting robust features from face ``appearance'' images in the infrared spectrum. They described a system which uses the wavelet transform based on a bank of Gabor filters. The marginal density functions of the filtered features are then modelled using \emph{Bessel K forms} which are matched using the simple $L_2$-norm. Srivastava \textit{et al.}\ reported a remarkable fit between the observed and the estimated marginals across a large set of filtered images. Evaluated on the Equinox database their method achieved a nearly perfect recognition rate and on the FSU database (the two databases are described in Sec.~\ref{ss:equinox} and~\ref{ss:fsu}) outperformed both Eigenfaces and independent component analysis based matching. A similar approach was also described by Buddharaju \textit{et al.}\ \cite{BuddPavlKaka2004}. The method of Nicolo and Schmid \cite{NicoSchm2011} also adopts Gabor wavelet features at its core and encodes the responses using the recently introduced Weber local descriptor \cite{ChenShanHeZhao+2010} and local binary patterns.

\subsubsection{Curvelet Transform}\label{sss:curvelet}
The curvelet transform an extension of the wavelet transform in which the degree of orientational localization is dependent on the scale of the curvelet \cite{MandMajuWu2007}. For a variety of natural images, the curvelet transform facilitates a sparser representation than wavelet transforms do, with effective spatial and directional localization of edge-like structures. Xie \textit{et al.}\ \cite{XieWuLiuFang2009a,XieWuLiuFang2009b,XieLiuWuLu2009c} described the first infrared based face recognition system which uses the curvelet transform for feature extraction. Using a simple nearest neighbour classifier, in their experiments the method demonstrated a slight advantage (of approximately 1-2\%) over simple linear discriminant based approaches, but with a significant improvement in computational and storage demands.

\subsubsection{Vascular Networks}\label{sss:vascular}
\begin{figure}[htb]
  \centering
  \small Vascular network of Buddharaju \textit{et al.}\\
  \subfigure[100\%]{\includegraphics[width=0.10\textwidth]{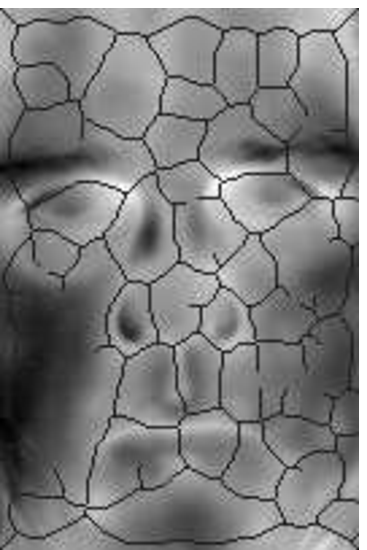}}~~
  \subfigure[ 90\%]{\includegraphics[width=0.10\textwidth]{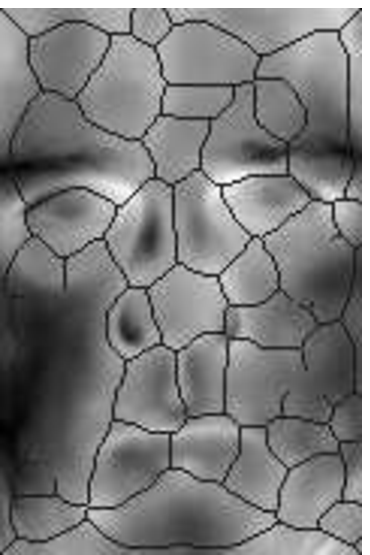}}~~
  \subfigure[ 80\%]{\includegraphics[width=0.10\textwidth]{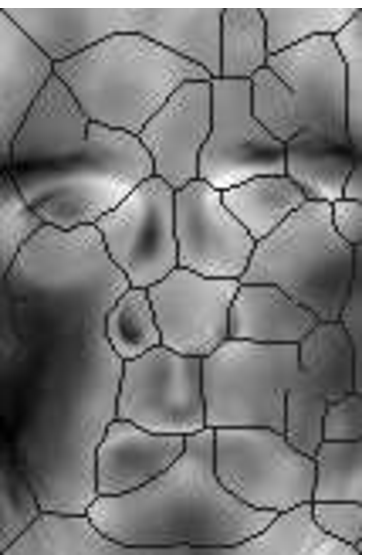}}~~
  \subfigure[ 70\%]{\includegraphics[width=0.10\textwidth]{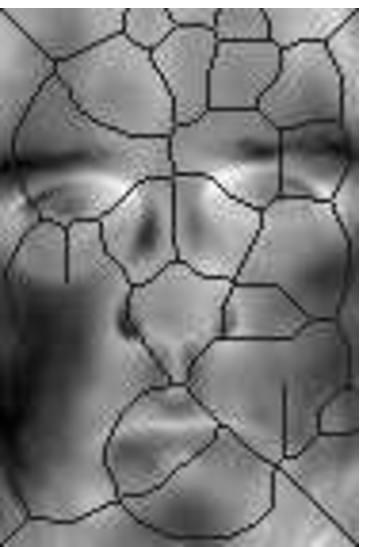}}
  \\
  \small Vesselness response based representation of Ghiass \textit{et al.}\\
  \subfigure[100\%]{\includegraphics[width=0.10\textwidth]{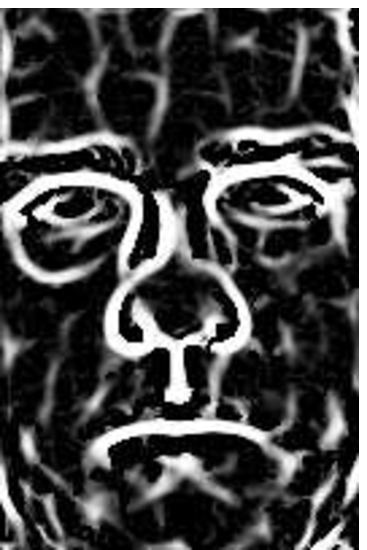}}~~
  \subfigure[ 90\%]{\includegraphics[width=0.10\textwidth]{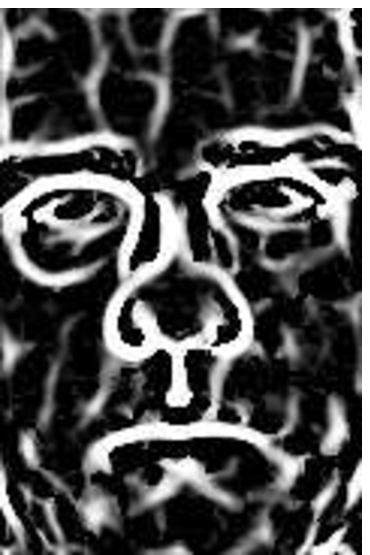}}~~
  \subfigure[ 80\%]{\includegraphics[width=0.10\textwidth]{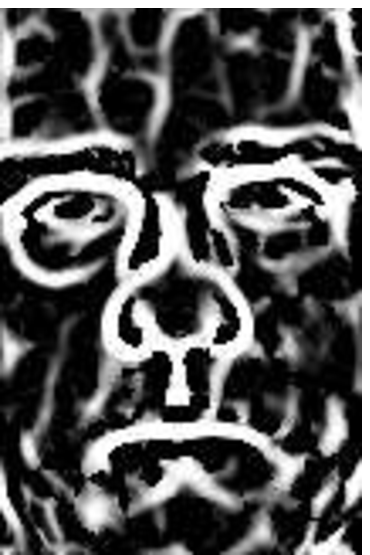}}~~
  \subfigure[ 70\%]{\includegraphics[width=0.10\textwidth]{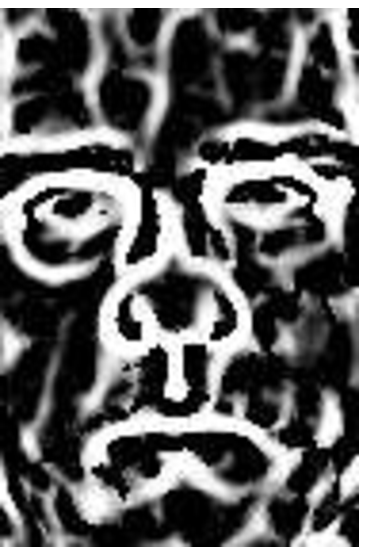}}
  \caption{ One of the major limitations of the vascular network based approach proposed by Buddharaju \textit{et al.} lies in its `crisp' binary nature: a particular pixel is deemed either a part of the vascular network or not. The consequence of this is that the extracted vascular network is highly sensitive to the scale of the input image (and thus to the distance of the user from the camera as well as the spatial resolution of the camera). (a-d) Even small changes in face scale can effect large topological changes on the result (note that the representation of interest is the vascular network, shown in black, which is only superimposed on the images it is extracted from for the benefit of the reader). (e-h) In contrast, the vesselness response based representation of Ghiass \textit{et al.}\ \cite{GhiaAranBendMald2013,GhiaAranBendMald2013a} encodes the certainty that a particular pixel locus is a reliable vessel pattern, and exhibits far greater resilience to scale changes.}
  \label{f:scale}
\end{figure}

Although the idea of using the superficial vascular network of a face to derive robust features for recognition dates as far back as the work of Prokoski \textit{et al.}\ \cite{ProkRiedCoff1992}, it wasn't until only recently that the first automatic methods have been described in the literature. The first corpus of work based around this idea was published by Buddharaju \textit{et al.}\ \cite{BuddPavlTsia2005,BuddPavlTsia2006,BuddPavlTsiaBaza2007} with subsequent further contributions by Gault \textit{et al.}\ \cite{GaulBlumFaraStar2010} and Seal \textit{et al.}\ \cite{SealNasiBhatBasu2011}. Following automatic background-foreground segmentation of a face, Buddharaju \textit{et al.}\ first extract blood vessels from an image using simple morphological filters, as shown in Fig.~\ref{f:scale}(a-d). The skeletonized vascular network is then used to localize salient features of the network which they term \emph{thermal minutia points} and which are similar in nature to the minutiae used in fingerprint recognition. Indeed, the authors adopt a method of matching sets of minutia points already widely used in fingerprint recognition, using relative minutiae orientations on local and global scale. Unsurprisingly, the method's performance was best when the semi-profile pose was used for training and querying, rather than the frontal pose. This finding is similar to what has repeatedly been noted by multiple authors for both human and computer based recognition in the visible spectrum \cite{SimZhang2004,LeeMoghPfisMach2004,AranCipo2004a}. While images of frontally oriented faces contain the highest degree of appearance redundancy, they limit the amount of discriminative information available from the sides of the face. In the multi-pose training scenario, rank-1 recognition of approximately 86\% and the equal error rate of approximately 18\% were achieved. While, as the authors note, some of the errors can be attributed to incorrectly localized thermal minutia points, the main reason for the relatively poor performance of their method is to be found in the sensitivity of their geometry based approach to out-of-plane rotation and the effected distortion of the observed vascular network shape.

In their more recent work, Buddharaju \textit{et al.}\ \cite{BuddPavl2009} improve their method on several accounts. Firstly, they introduce a post-processing step in their vascular network segmentation algorithm, with the aim of removing spurious segments which, as mentioned previously, are responsible for some of the matching errors observed of their initial method \cite{BuddPavlTsiaBaza2007}. More significantly, using an iterative closest point algorithm Buddharaju \textit{et al.}\ now also non-rigidly register two vascular networks which are being compared as a means of correcting for the distortion effected by out-of-plane head rotation. Their experiments indeed demonstrate the superiority of this approach over that proposed previously.

Cho \textit{et al.}\ \cite{ChoWangOng2009} describe a simple modification of the temporal minutia point based approach of Buddharaju \textit{et al.}\ which appends the location of the face centre (estimated from the segmented foreground mask) to the vectors corresponding to minutia point loci. Their method significantly outperformed Na\"{\i}ve Bayes, multilayer perceptron and Adaboost classifiers, achieving a false acceptance rate of 1.2\% for the false rejection rate of 0.1\% on the Equinox database (see Sec.~\ref{s:dbs}).

The most recent contribution to the corpus of work on vascular network based recognition was made by Ghiass \textit{et al.}\ \cite{GhiaAranBendMald2013,GhiaAranBendMald2013a}. There are several important aspects of novelty in the approach they describe. Firstly, instead of seeking a binary representation in which each pixel either `crisply' belongs or does not belong to the vascular network, the baseline representation of Ghiass \textit{et al.}\ smoothly encodes this membership by a confidence level in the interval $[0,1]$. This change of paradigm, further embedded within a multi-scale vascular network extraction framework, is shown to achieve better robustness to face scale changes (e.g.\ due to different resolutions of query and training images, or indeed different user-camera distances), as illustrated in Fig.~\ref{f:scale}. The second significant contribution of this work concerns the recognition across pose which is a major challenge for previously proposed vascular network based methods. The method of Ghiass \textit{et al.}\ achieves pose invariance by geometrically warping images to a canonical frame. Ghiass \textit{et al.}\ are the first to show how the active appearance model (AAM) \cite{CootEdwaTayl1998} can be applied on IR images of faces and, specifically, they show how the difficult problem of AAM convergence in the presence of many local minima can be addressed by pre-processing thermal IR images in a manner which emphasizes discriminative information content~\cite{GhiaAranBendMald2013}. In their most recent work, recognition across the entire range of poses from frontal to profile is achieved by training en ensemble of AAMs, each `specializing' in a particular region of the thermal IR face space corresponding to an automatically determined cluster of poses and subject appearances~\cite{GhiaAranBendMald2013a}.

Lastly, it should be noted that Ghiass \textit{et al.}\ emphasize that ``$\ldots$none of the existing publications on face recognition using `vascular network' based representations provide any evidence that the extracted structures are indeed blood vessels. Thus the reader should understand that we use this term for the sake of consistency with previous work, and that we do \emph{not} claim that what we extract in this paper is an actual vascular network. Rather we prefer to think of our representation as a \emph{function} of the underlying vasculature'' (the reader may also find the work of Gault \textit{et al.}\ \cite{GaulBlumFaraStar2010} useful in the consideration of this issue).

\subsubsection{Blood Perfusion}\label{sss:perfusion}
\begin{figure}[htb]
  \centering
  \subfigure[Thermogram]{\includegraphics[height=0.25\textwidth]{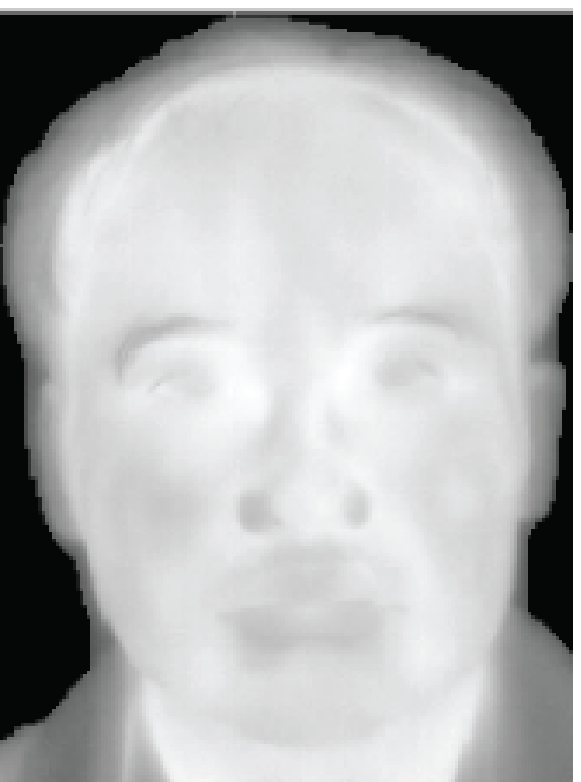}}\hspace{30pt}
  \subfigure[Perfusion]{\includegraphics[height=0.25\textwidth]{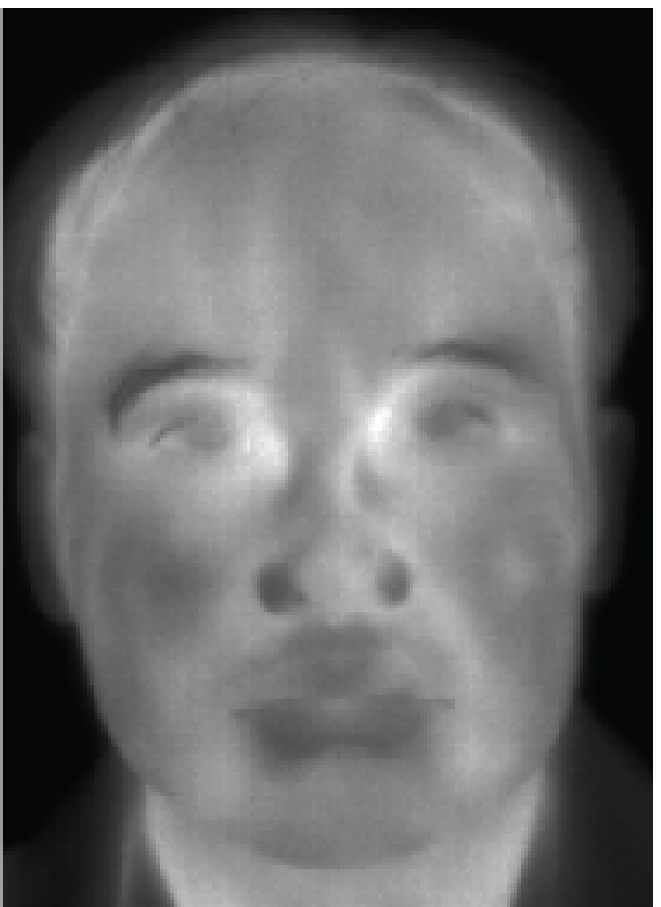}}
  \caption{ (a) A thermogram and the corresponding (b) blood perfusion image.  }
  \label{f:bloodP}
\end{figure}

A different attempt at extracting invariant features which also exploits the temperature differential between vascular and non-vascular tissues was proposed by Wu \textit{et al.}\ \cite{WuSongJianXie+2005} and Xie \textit{et al.}\ \cite{XieLiuWuFang2009}. Using a series of assumptions on relative temperatures of body's deep and superficial tissues, and the ambient temperature, Wu \textit{et al.}\ formulate a differential equation governing blood perfusion. The model is then used to compute a ``blood perfusion image'' from the original segmented thermogram of a face, as illustrated in Fig.~\ref{f:bloodP}. Finally, blood perfusion images are matched using a standard linear discriminant and an RBF network.

Following their original work, Wu \textit{et al.}\ \cite{WuGuChiaOng2007a} and Xie \textit{et al.}\ \cite{XieWuHeFang+2010} introduce alternative blood perfusion models. The model described by Wu \textit{et al.}\ was demonstrated to produce comparable recognition results to the more complex model previously, while achieving greater time and storage efficiency. Xie \textit{et al.}\ derived a model based on the Pennes equation which too outperformed the initial model described by Wu \textit{et al.}\ \cite{WuSongJianXie+2005}.

In addition to their work on different blood perfusion models, in their more recent work Wu \textit{et al.}\ \cite{WuWeiFangLi+2007} also extend their classification method by another feature extraction stage. Instead of using the blood perfusion image directly, they first decompose the image of a face using the wavelet transform. After that, they apply the sub-block discrete cosine transform on the low frequency sub-band of the transform and use the obtained coefficients as an identity descriptor. Wu \textit{et al.}\ demonstrate experimentally that this representation outperforms both purely discrete cosine transform based and purely wavelet transform based representations of the blood perfusion image.

\subsection{Multi-Spectral and Hyper-Spectral Methods}\label{ss:spectral}
Multi-spectral imaging refers to the process of concurrent acquisition of a set of images, each image corresponding to a different band of the electromagnetic spectrum. A familiar example is colour imaging in the visual spectrum which acquires three images that correspond to what the human eyes perceives as red, green, and blue sensations. In general, the number of bands can be much greater and the width of the sub-bands different images correspond to wider or narrower. The terms multi-spectral and hyper-spectral imaging are often used interchangeably, while some authors make the distinction between sets of images acquired in discrete and separated narrow bands (multi-spectral) and sets of images acquired in usually wider but frequency wise contiguous sub-bands. Henceforth in this paper we will consistently use the term multi-spectral imaging and specifically describe the data used by a specific method (or reference a standard database which contains this information).

\begin{figure}[htb]
  \centering
  \footnotesize
  \begin{tabular}{VVV}
      \includegraphics[width=0.11\textwidth]{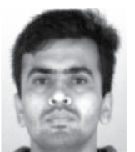}&
      \includegraphics[width=0.11\textwidth]{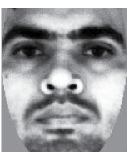}&
      \begin{tabular}{ccccc}
        \includegraphics[width=0.11\textwidth]{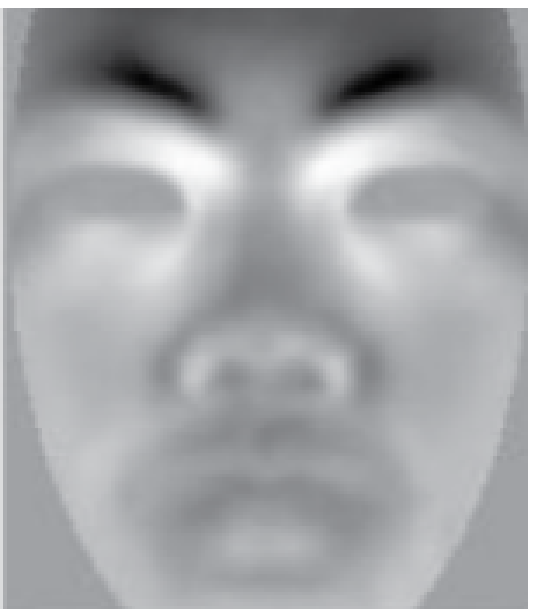}&
        \includegraphics[width=0.11\textwidth]{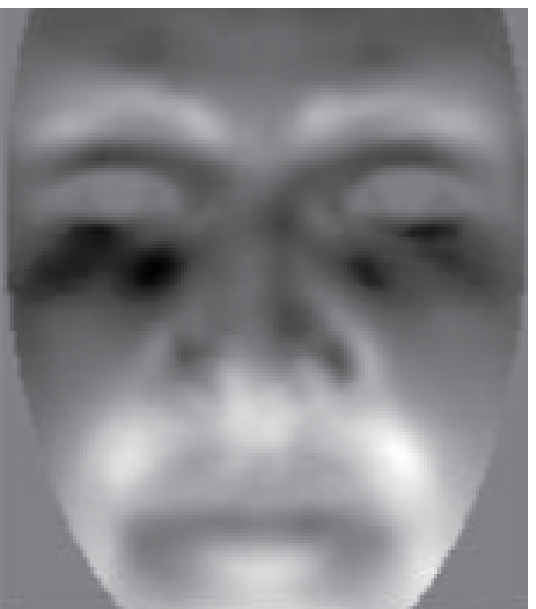}&
        \includegraphics[width=0.11\textwidth]{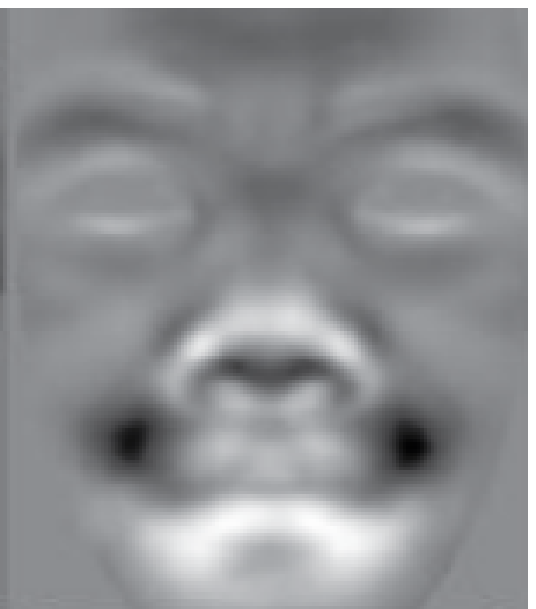}&
        \includegraphics[width=0.11\textwidth]{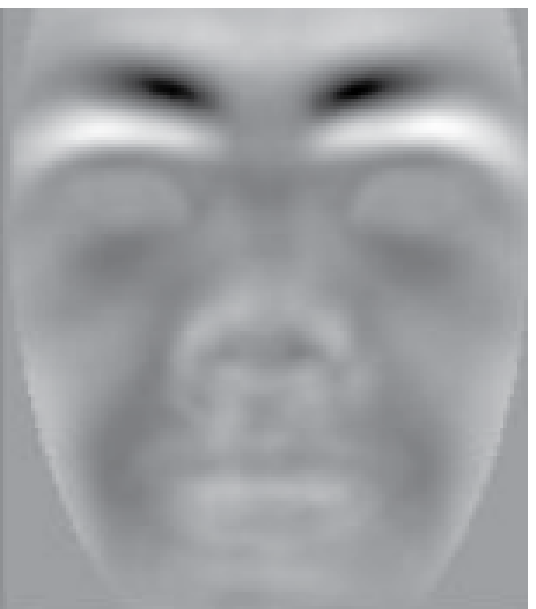}&
        \includegraphics[width=0.11\textwidth]{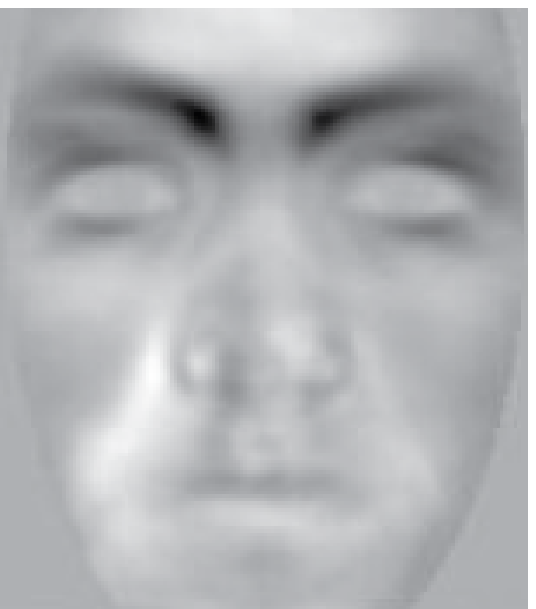}
      \end{tabular}\\
      (a) & (b) & (c)
  \end{tabular}
  \caption{ (a) The original visible spectrum image, (b) the corresponding spectral-face, and (c) the first five eigen-spectral-faces obtained by Pan \textit{et al.}\
            \cite{PanHealPrasTrom2005}. }
  \label{f:spectralFace}
\end{figure}

The epidermal and dermal layers of skin make up a scattering medium that contains pigments such as melanin, hemoglobin, bilirubin, and $\beta$-carotene. Small changes in the distribution of these pigments induce significant changes in the skin's spectral reflectance. In the method of Pan \textit{et al.}\ \cite{PanHealPrasTrom2003}, the structure of the skin, including sub-surface layers, is sensed using multi-spectral imaging in 31 narrow bands of the NIR sub-band. The authors measured the variability in spectral properties of the human skin and showed that there are significant differences in both amplitude and spectral shape of the reflectance curves for the different subjects, while the spectral reflectance for the same subject did not change in different trials. They also observed good invariance of local spectral properties to face orientation and expression. On a proprietary database of 200 subjects with a diverse sex, age and ethnicity composition, the proposed method achieved recognition rates of 50\%, 75\%, and 92\% for profile, semi-profile and frontal faces respectively. In their subsequent work, Pan \textit{et al.}\ \cite{PanHealPrasTrom2005} examine the use of holistic multi-spectral appearance, in contrast to their previous work which used a sparse set of local features only. They apply Eigenfaces on images obtained from different NIR sub-bands, as a means of de-correlating the set of features used for classification. They also describe a method for synthesizing a discriminative signature image that they term the ``spectral-face'' image, obtained by sequential interlacing of images corresponding to different sub-bands, which in their experiments showed some advantage when used as input for Eigenfaces. An example of a spectral-face image and spectral-face based eigenfaces is shown in Fig.~\ref{f:spectralFace}.

\subsubsection{Inter-Spectral Matching}\label{sss:interspectral}
The work by Bourlai \textit{et al.}\ \cite{BourKalkRossCuki+2010} is the only published account of the use of data acquired in the short wave infrared sub-band for face recognition. Following face localization using the detector of Viola and Jones \cite{ViolJone2004}, Bourlai \textit{et al.}\ apply contrast limited adaptive histogram equalization and feed the result into: (i) a $K$-nearest neighbour based classifier, (ii) VeriLook's and (iii) Identity Tools G8 commercial recognition systems. A particularly interesting aspect of this work is that Bourlai \textit{et al.}\ investigate the possibility of inter-spectral matching. Their experimental results suggest that SWIR images can be matched to visible images with promising results. Klare and Jain \cite{KlarJain2010} similarly match visible and NIR data, using local binary patterns and HoG local descriptors \cite{DalaTrig2005}. The success of these methods not particularly surprising considering that the NIR and SWIR sub-bands of the infrared spectrum is much closer to the visible spectrum than MWIR or LWIR sub-bands. Indeed, this premise is central to the methods described by Chen \textit{et al.}\ \cite{ChenYiYangZhao+2009}, Lei and Li \cite{LeiLi2009}, Mavadati \textit{et al.}\ \cite{MavaSadeKitt2010} and Shao \textit{et al.}\ \cite{ShaoWangWang2009} who show that visible spectrum data can be used to create synthetic NIR images, the NIR sub-band of the infrared being the closest to the visible spectrum.

A greater challenge was recently investigated by Bourlai \textit{et al.} \cite{BourRossChenHorn2012} who attempted to match MWIR to visible spectrum images. Following global affine normalization and contrast limited adaptive histogram equalization, the authors evaluated different pre-processing methods (the self-quotient image and difference of Gaussian based filtering), feature types (local binary patterns, pyramids of oriented gradients histograms \cite{BoscZissMuno2007} and scale invariant feature transform \cite{Lowe2004}) and similarity measures (chi-squared, distance transform based, Euclidean and city-block). No combination of the parameters was found to be very promising, the best performing patch based and difference of Gaussian filtered LBP on average achieving only approximately 40\% correct rank-1 recognition rate on a 39 subject subset of the West Virginia University database (see Sec.~\ref{ss:WVUM}).

\subsection{Multimodal Methods}\label{ss:multimodal}
As predicted from theory and repeatedly demonstrated in experiments summarized in the preceding sections, some of the major challenges of automatic face recognition methods which use infrared images include the opaqueness of eyeglasses in this spectrum and the dependence of the acquired data on the emotional and physical condition of the subject. In contrast, neither of these is a significant challenge in the visible spectrum. In the visible spectrum eyeglasses are largely transparent and such physiological variables such as the emotional state have negligible inherent effect on one's appearance. Indeed, in the context of many challenging factors in the two spectra, they can be considered complementary. Consequently, it can be expected that this complementary information can be exploited to achieve a greater degree of invariance across a wide range of nuisance variables.

Most of the methods for fusing information from visible and infrared spectra described in the literature fall into one of two groups. The first of these is data-level fusion. Methods of this category construct features which inherit information from both modalities, and then perform learning and classification of such features. The second fusion type is decision-level. Methods of this group compute the final score of matching two individuals from matches independently performed in the visible and in the infrared spectra. To date, decision-level fusion predominates in the infrared face recognition literature.

\subsubsection{Early Work}
Wilder \textit{et al.}\ \cite{WildPhilJianWien1996} were the first to investigate the possibility of fusion of visible and infrared data. They examined three different methods for representing and matching images, using (i) transform coded grey scale projections, (ii) Eigenfaces and (iii) pursuit filters, and compared the performance of the two modalities in isolation and their fusion. Decision-level fusion was achieved simply by adding the matching scores separately computed for visible and infrared data. The transform coded grey scale projections based method achieved the best performance of the three methods compared. Using this representation independently in the visible and thermal IR spectra, the two modalities achieved comparable recognition results. However, the proposed fusion method had a remarkable effect, reducing the error rate for approximately an order of magnitude (from approximately 10\% down to approximately 1\%).


\subsubsection{Time-Lapse}
The problem of time-lapse in recognition concerns the empirical observation made across different recognition methodologies that the performance of an algorithm degrades with the passage of time between training and test data even if the acquisition conditions are seemingly the same. The term ``time-lapse'' is, we would argue, a somewhat misleading one. Clearly, the drop in recognition performance is not caused by the passage of time \textit{per se} but rather a change in some tangible factor which affects facial appearance. This is particularly easy to illustrate on thermal data. Even if external imaging conditions are controlled or compensated for, none of the published work attempts to control or measure the effects of the emotional state or the level of excitement of the subject\footnote{This could be achieved using various proxy variables correlated with sympathetic nervous system output, for example, such as perspiration rate, pulse, galvanic skin response and so on.} or indeed the loss of calibration of the infrared camera \cite{Mald2001}. The effects of external temperature on the temperature of the face is explicitly handled only in the method proposed by Siddiqui \textit{et al.}\ \cite{SiddSherKhal2004} who used simple thresholding and image enhancement to detect and normalize the appearance of face regions with particularly delayed temperature regulation. Nonetheless, for the sake of consistency and uniformity with the rest of the literature, we shall continue using the term ``time-lapse'' with an implicit understanding of the underlying issues raised herein.

The effect of time-lapse on the performance of infrared based systems was investigated by Chen \textit{et al.}\ \cite{ChenFlynBowy2003,ChenFlynBowy2003a,ChenFlynBowy2005}. They presented experiments evidencing the complementarity of visible and infrared spectra in the presence of time-lapse by showing that recognition errors achieved using the two modalities, and effected by the passage of time between training and query data acquisition, are largely uncorrelated. Similar observations were made by Socolinsky \textit{et al.}\ \cite{SocoSeli2004}. Regardless of whether simple PCA features were used for matching or the commercial system developed by the Equinox Corporation, the benefit of fusing visible and infrared modalities was substantial even though the simple additive combination of matching scores was used.

\subsubsection{Eyeglasses}\label{sss:glasses}
Since eyeglasses are opaque to the infrared frequencies in the SWIR, MWIR and LWIR sub-bands \cite{TasmJaeg2009}, their presence is a major issue when this data is used for recognition as some of the most discriminative regions of the face can be occluded. In contrast, the effect of eyeglasses on the appearance in the visible spectrum is far less significant. The methods of Gyaourova \textit{et al.}\ \cite{GyaoBebiPavl2004} and Singh \textit{et al.}\ \cite{SingGyaoBebi+2004} propose a data level fusion approach whereby a genetic algorithm is used to select features computed separately in the visual and thermal infrared spectra. Using two types of features, Haar wavelet based and eigencomponent based, and the Equinox database the proposed fusion method was shown to yield a superior performance compared to both purely visual and purely thermal infrared based matching, and particularly so in the presence of eyeglasses or variable illumination. Inspired by this work, Chen \textit{et al.}\ \cite{ChenJingXiao2005a} describe a similar fusion method. Instead of a genetic algorithm, they employ a fuzzy integral neural network based feature selection algorithm which has the advantage of faster convergence and greater probability of reaching a solution close to the global optimum.

Heo \textit{et al.}\ \cite{HeoKongAbidAbid2004} investigate both data-level and decision-level fusion. First, following the detection of eyeglasses, the corresponding image region is replaced with a generic eye template. As expected, the replacement of the eyeglass region with a generic template significantly improves recognition in the thermal but not in the visible spectrum. Data-level fusion is achieved by simple weighted addition of the corresponding pixels in mutually co-registered visible and thermal infrared images. The key contribution of this work pertains to the difference in performance observed between data-level and decision-level fusion. Interestingly, unlike in the case of data-level fusion where a remarkable performance improvement was observed, when fusion was performed at the decision-level the performance was actually somewhat worsened.

A similar approach to handing the occlusion of thermal infrared image regions by eyeglasses was taken by Kong \textit{et al.}\ \cite{KongHeoBougZheng+2007}. They replace an elliptical patch surrounding the eye occluded by eyeglasses with a patch representing the average eye appearance. Although differently implemented, the approach of Arandjelovi\'c \textit{et al.}\ \cite{AranHammCipo2010} is similar in spirit. Following the detection of eyeglasses unlike Heo \textit{et al.}\ and Kong \textit{et al.}\ Arandjelovi\'c \textit{et al.}\ do not remove the offending image region, but rather introduce a robust modification to canonical correlations based matching, which ignores the eyeglasses region when sets of images are compared.

\subsubsection{Illumination}
In addition to the problem posed by eyeglasses, in their work already described in Sec.~\ref{sss:glasses} Heo \textit{et al.}\ \cite{HeoKongAbidAbid2004} also examined the effects of the proposed fusion on illumination invariance. Their results successfully substantiated the theoretically expected complementarity of infrared and visible spectrum data.

Socolinsky \textit{et al.}\ \cite{SocoSeliNeuh2003,SocoSeli2004a} extend their previous work \cite{SocoSeli2002} by describing a simple decision based fusion based on a weighted combination of visible and thermal infrared based matching scores, and evaluate it in indoor and outdoor data acquisition environments. The more extreme illumination conditions encountered outdoors proved rather more challenging than the indoor environment, regardless of which modality or baseline matching algorithm was used for recognition. Although simple, their fusion approach did yield substantial improvements in all cases, but still failed reach practically useful performance levels when applied outdoors.

In spirit, the work of Bhowmik \textit{et al.}\ \cite{BhowBhatNasiBasu+2010} builds on the contribution of Socolinsky \textit{et al.}\ \cite{SocoSeliNeuh2003,SocoSeli2004a}. Bhowmik \textit{et al.}\ also investigate a simple weighted combination of visible and thermal infrared spectrum matching scores and report the performance of the fusion for different contributions of the two.

\begin{figure}[htb]
  \centering
  \includegraphics[height=0.35\textwidth]{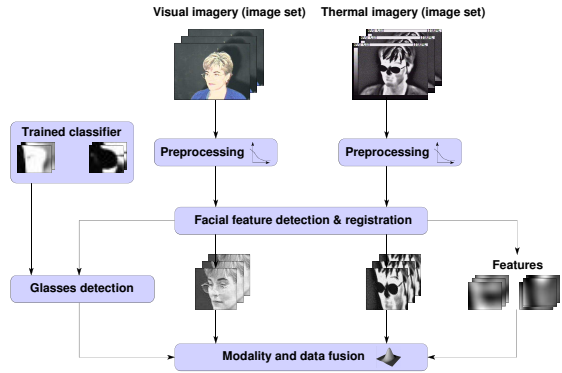}
  \caption{ The method proposed by Arandjelovi\'c \textit{et al.}\
            \cite{AranHammCipo2006,AranHammCipo2010} comprises (i) data preprocessing and registration, (ii) glasses detection and (iii) fusion of holistic and local face representations using visual and thermal modalities. }
  \label{f:aranOverview}
\end{figure}

The limitations of the approaches of Socolinsky \textit{et al.}\ and Bhowmik \textit{et al.}\ was recognized by Arandjelovi\'c \textit{et al.}\ \cite{AranHammCipo2010}, who demonstrate that the optimal weights in decision-level fusion are illumination dependent.
In a series of works Arandjelovi\'c \textit{et al.}\ \cite{AranHammCipo2006,AranHammCipo2006a,AranHammCipo2010} extend their method aimed at achieving illumination invariance using visible spectrum data only \cite{AranCipo2006a}, which fused raw appearance and filtered appearance based matching scores, and apply it to the fusion of matching scores based on visible and thermal data. A block diagram of their system is illustrated in Fig.~\ref{f:aranOverview}. Their main contribution is a fusion method which learns the optimal weighting of matching scores in an illumination-specific manner. Illumination specificity is achieved implicitly. Conceptually, they exploit the observation that if the best match in the visible domain is sufficiently confident, the illumination change between training and novel data is small so more weight should be placed on the visible spectrum match. If the best match is insufficiently confident, the illumination change is significant and more weight is placed on infrared data which is largely unaffected by visible light. Conceptually similar is the fusion approach described by Moon \textit{et al.}\ \cite{MoonKongYooChun2006} which also adaptively controls the contributions of the visible and thermal infrared spectra. Unlike the Arandjelovi\'c \textit{et al.}\ who use a combination of filtered holistic and local appearances, Moon \textit{et al.}\ represent images of faces using the coefficients obtained from a wavelet decomposition of an input image. Different wavelet based fusion approaches have also been proposed by Kwon \textit{et al.}\ \cite{KwonKong2005} and Zahran \textit{et al.}\ \cite{ZahrAbbaDessAsho+2009}.

\subsubsection{Expression}
The method proposed by Hariharan \textit{et al.}\ \cite{HariKoscAbidGrib+2006} is one of the small number data-level fusion approaches. Hariharan \textit{et al.}\ produce a synthetic image which contains information from both visible and infrared spectra. The key element of their approach is empirical mode decomposition. After decomposing the corresponding and mutually co-registered visible and thermal infrared spectrum images into their intrinsic mode functions, a new image is produced as a re-weighted sum of the intrinsic mode functions of both modalities. The re-weighting coefficients are determined experimentally on a training set in an \textit{ad hoc} subjective manner which involves human judgement on how discriminative the resulting image appears. Hariharan \textit{et al.}\ report that their method outperformed that proposed by Kong \textit{et al.}\ \cite{KongHeoBougZheng+2007}, as well as Rockinger and Fechner \cite{RockFech1998}, and particularly so in poor illumination conditions and in the presence of facial expression changes.

\subsection{Other Approaches}
Owing to the increasing popularity of research into infrared based recognition there are a number of approaches in the literature which we did not discuss explicitly. These include the geometric invariant moment based approaches of Abas and Ono \cite{AbasOno2009,AbasOno2009a,AbasOno2010}, elastic graph matching based method of Hizem textit{et al.}\ \cite{HizeAllaMellDori2009}, isotherm based method of Tzeng \textit{et al.}\ \cite{TzenLeeChen2011}, faceprints of Akhloufi and Bendada \cite{AkhlBend2008}, fusion work of Toh \textit{et al.}\ \cite{TohKimLee2008,Toh2010}, and others \cite{ShaoWang2009,ZahrAbbaDessAsho+2009a,AkhlBend2010,PopGordFlorVlai2010,AkhlBend2010a}. Specifically, we did not describe (i) straightforward or minor extensions of the original approaches already surveyed and (ii) those methods which lack the weight of sufficient empirical evidence to support their competitiveness with the state-of-the-art at the time when they were first proposed. Nonetheless references to these are provided herein for the sake of completeness and for the benefit of the reader.

\section{Infrared Face Databases}\label{s:dbs}
The previous section makes it readily apparent that a major obstacle to understanding relative merits of published work on infrared based face recognition lies in the evaluation methodology used to assess the effectiveness of proposed approaches. Different authors focus their attention to different nuisance variables and, in the best case, evaluate their method on appropriate data sets. However, it is largely unclear, at least on the basis of empirical evidence, how different methods compare to one another if they are evaluated on the same data representative of that which may be acquired in a real-world application. In this section we review the most relevant databases of infrared imagery which have been collected for research purposes. We focus our attention on those which are public, that is, freely available. A quick reference summary of the key facts can be found in Table~\ref{t:dbSummary}.

\begin{table}
  \caption{ A quick reference summary of the main databases of face images acquired in the infrared spectrum. The presence of variability due to a particular nuisance variable in the data is denoted by $\CIRCLE$, some but limited variability by $\RIGHTcircle$ and little to no variability by $\Circle$. }
  \centering
  \includegraphics[width=1.00\textwidth]{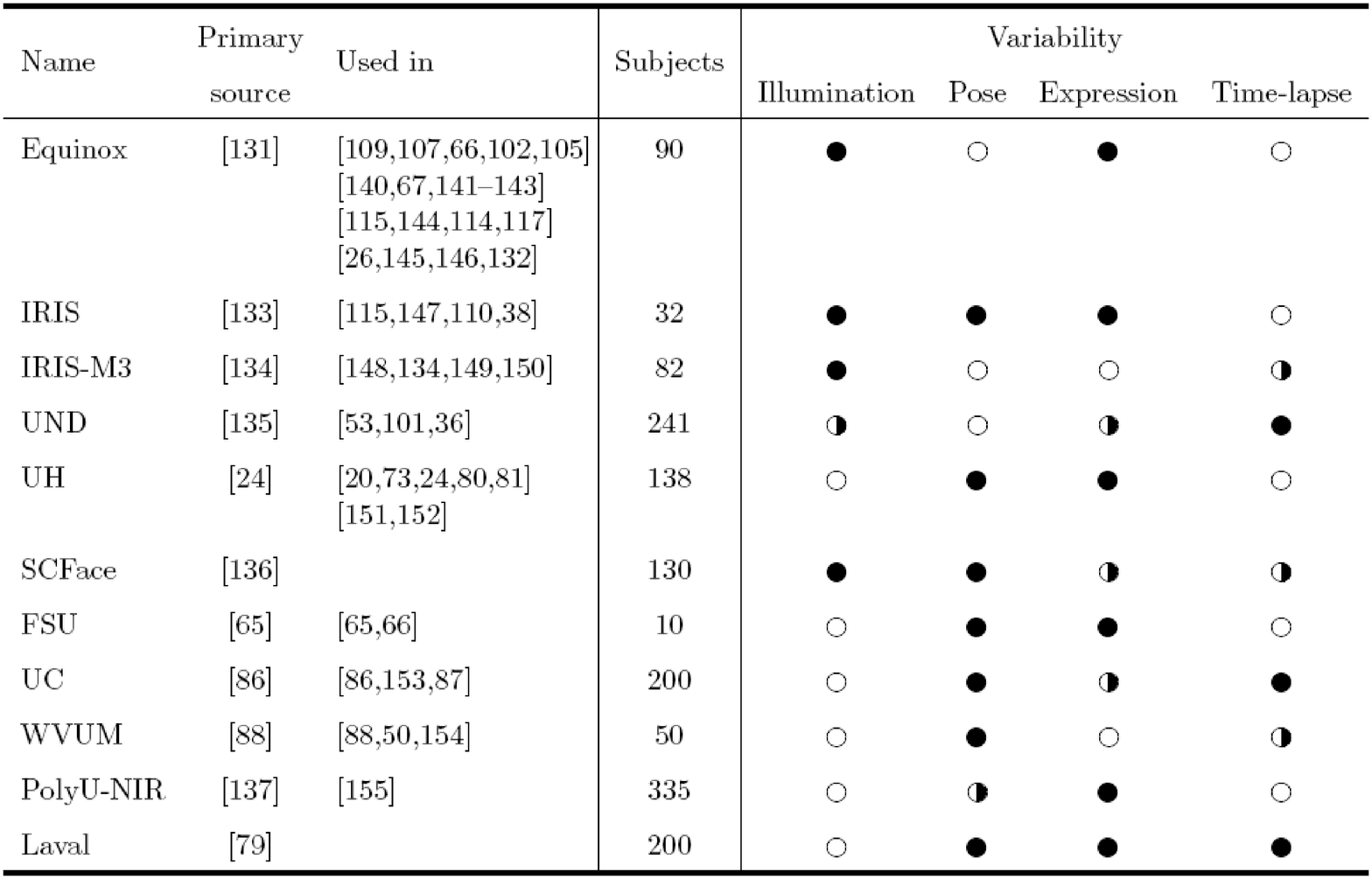}
  \label{t:dbSummary}
\end{table}

\subsection{Equinox}\label{ss:equinox}
\begin{figure}[htb]
  \centering
  \subfigure[]{\hspace{15pt}\includegraphics[width=0.15\textwidth]{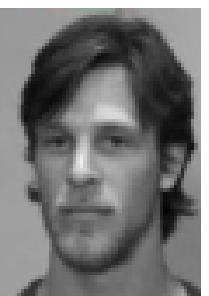}\hspace{15pt}}
  \subfigure[]{\hspace{15pt}\includegraphics[width=0.15\textwidth]{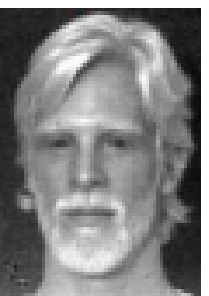}\hspace{15pt}}
  \subfigure[]{\hspace{15pt}\includegraphics[width=0.15\textwidth]{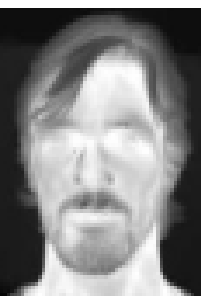}\hspace{15pt}}
  \subfigure[]{\hspace{15pt}\includegraphics[width=0.15\textwidth]{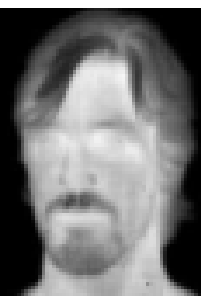}\hspace{15pt}}
  \caption{ Four concurrently acquired images from the Equinox's ``Human Identification at a Distance Database'' respectively in the visible, long wave infrared,  medium wave
            infrared and short wave infrared spectral bands. Images are co-registered to within $1/3$ of a pixel. }
  \label{f:dbEquinox}
\end{figure}

The ``Human Identification at a Distance'' database \cite{DbEqui}, collected by Equinox Corporation has been the most used data set for the evaluation of infrared based face recognition algorithms in the literature. It is freely available for non-commercial use. The data set contains $240 \times 320$ pixel images of 90 individuals' appearance in the (i) visible, (ii) long wave infrared, (iii) medium wave infrared, and (iv) short wave infrared spectral bands, acquired using a setup of cameras co-registered to within $1/3$ of a pixel. Fig.~\ref{f:dbEquinox} shows an example of a set of four concurrently acquired images. For each subject in the database, data was collected under three different controlled lighting conditions using a directional light source illuminating from the (i) frontal, (ii) left lateral and (iii) right lateral directions. In all cases the subject was facing the camera so the database contains only frontal face images. Individuals wearing glasses were imaged with glasses both on and off. Facial expression variability was introduced by two means. First, a 4~second video sequence acquired at 10~fps was taken of the subject pronouncing the vowels. In addition, the subject was explicitly asked to assume the `smiling', `frowning' and `surprised' expressions.
Note that all images of a particular individual were acquired in a single session making this data set unsuitable for the evaluation of robustness to time-lapse associated appearance changes. A comprehensive evaluation of different recognition approaches on the Equinox database was published by Hermosilla \textit{et al.}\ \cite{HermRuizVersCorr2009}.

\subsection{IRIS Thermal/Visible}\label{ss:iris}
\begin{figure}[htb]
  \centering
  \includegraphics[width=1.00\textwidth]{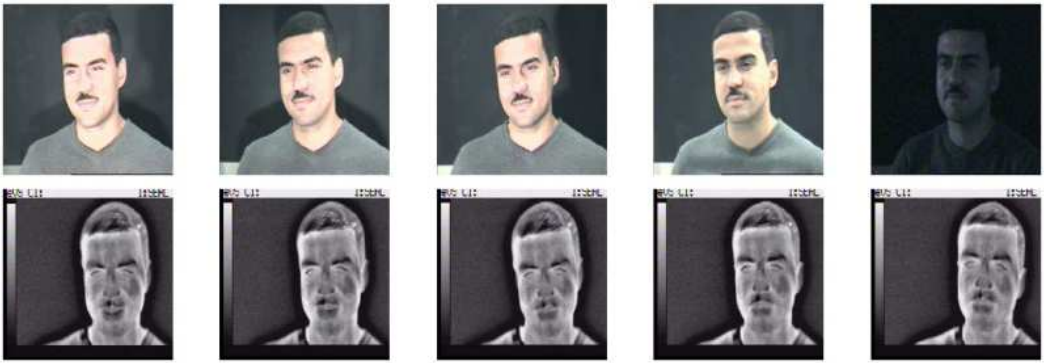}
  \caption{ Five pairs of matching visible (top row) and thermal (bottom) row images of the IRIS Thermal/Visible Face Database \cite{DbIRIS} database of a subject in the same
            pose and different illumination conditions. Note that the visible and thermal spectrum images are not mutually co-registered. }
  \label{f:dbIRIS}
\end{figure}

IRIS Thermal/Visible Face Database \cite{DbIRIS} is a free data set of thermal and visual spectrum images, collected across pose, illumination and expression variation. The set comprises 4228 pairs of $320 \times 240$ pixel images which were concurrently acquired but are not mutually co-registered. There are 32 individuals in the database, with 176--250 images per person. The five illumination conditions were obtained using different on/off combinations of two directional lateral light sources and one ambient light source: (i) all light sources off, (ii) only the ambient light on, (iii) the ambient and the left directional light on, (iv) the ambient and the right directional light on, and (v) all light on. In a similar manner as in the Equinox database, images of the subject `smiling', `frowning' and exhibiting `surprise' were acquired. Using a motorized setup, the camera viewing direction was controlled and images acquired every 36$^{\circ}$ across the 180$^{\circ}$ range, resulting in 11 images per modality for each illumination setting and subject expression. All data for a particular subject was acquired in a single session. Fig.~\ref{f:dbIRIS} shows examples of images from the database.

\subsection{IRIS-M3}\label{ss:m3}
Much like the Equinox and IRIS Thermal/Visible data sets, IRIS-M3~\cite{ChanHariYiKosc+2006a} is a database which contains both thermal and visible spectrum images. Unlike the previous two databases, it also includes multi-spectral images acquired in 25 sub-bands of the visible spectrum. The acquisition of multi spectral images was achieved using an electronically tunable liquid crystal filter coupled to a camera.

\begin{figure}[htb]
  \centering
  \subfigure[Flourescent]{\includegraphics[width=0.22\textwidth]{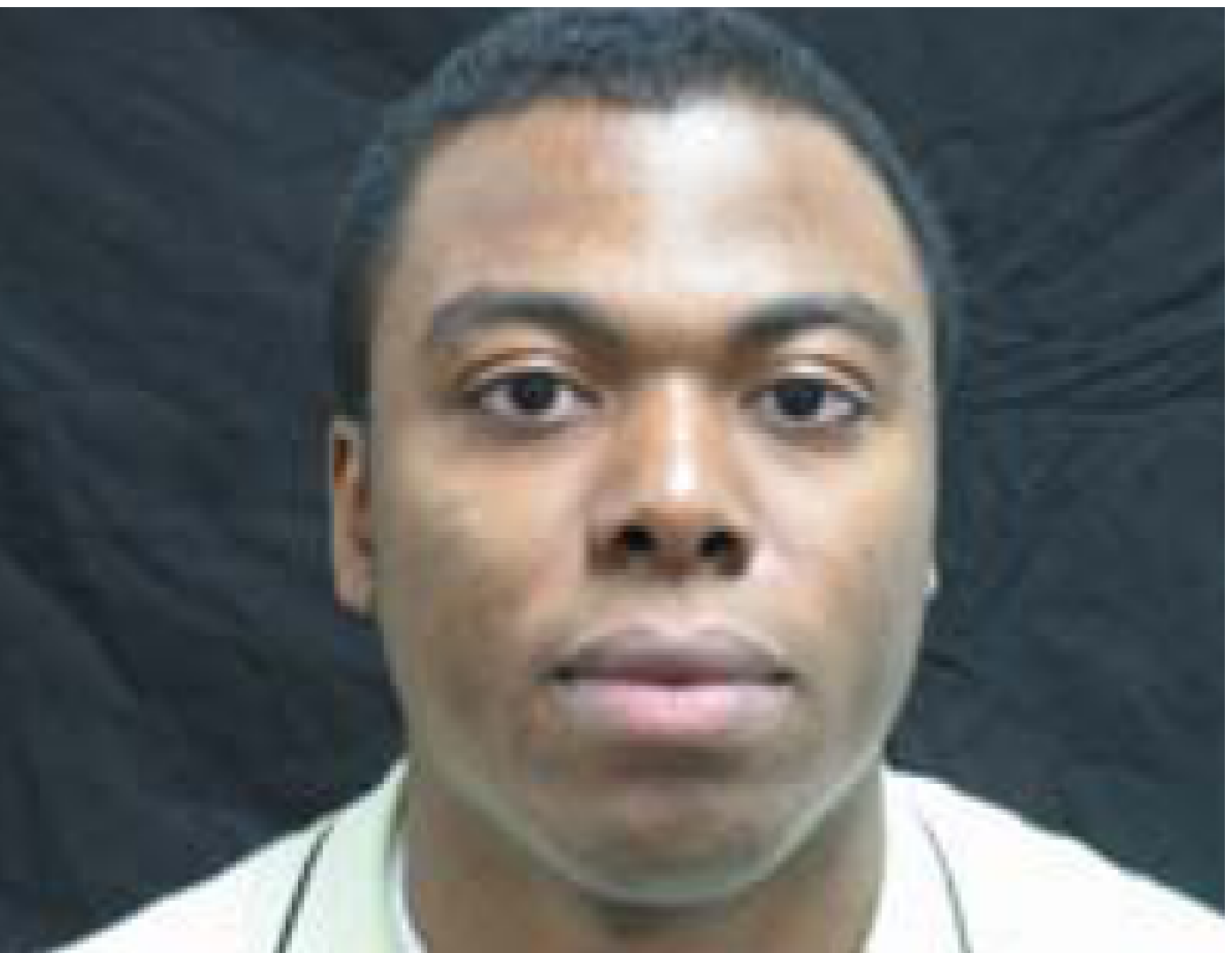}}
  \subfigure[Thermal]{\includegraphics[width=0.22\textwidth]{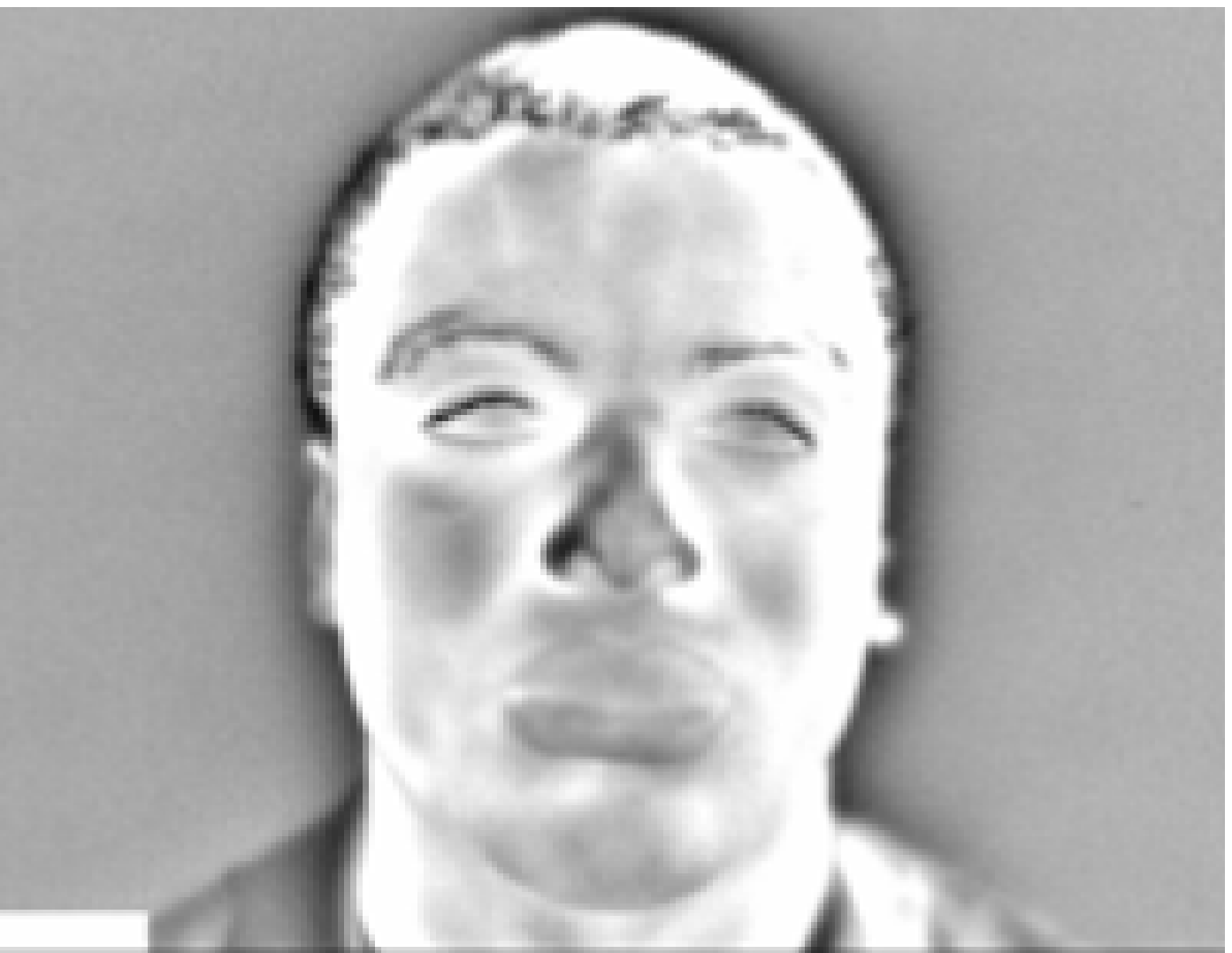}}
  \subfigure[480nm]{\includegraphics[width=0.22\textwidth]{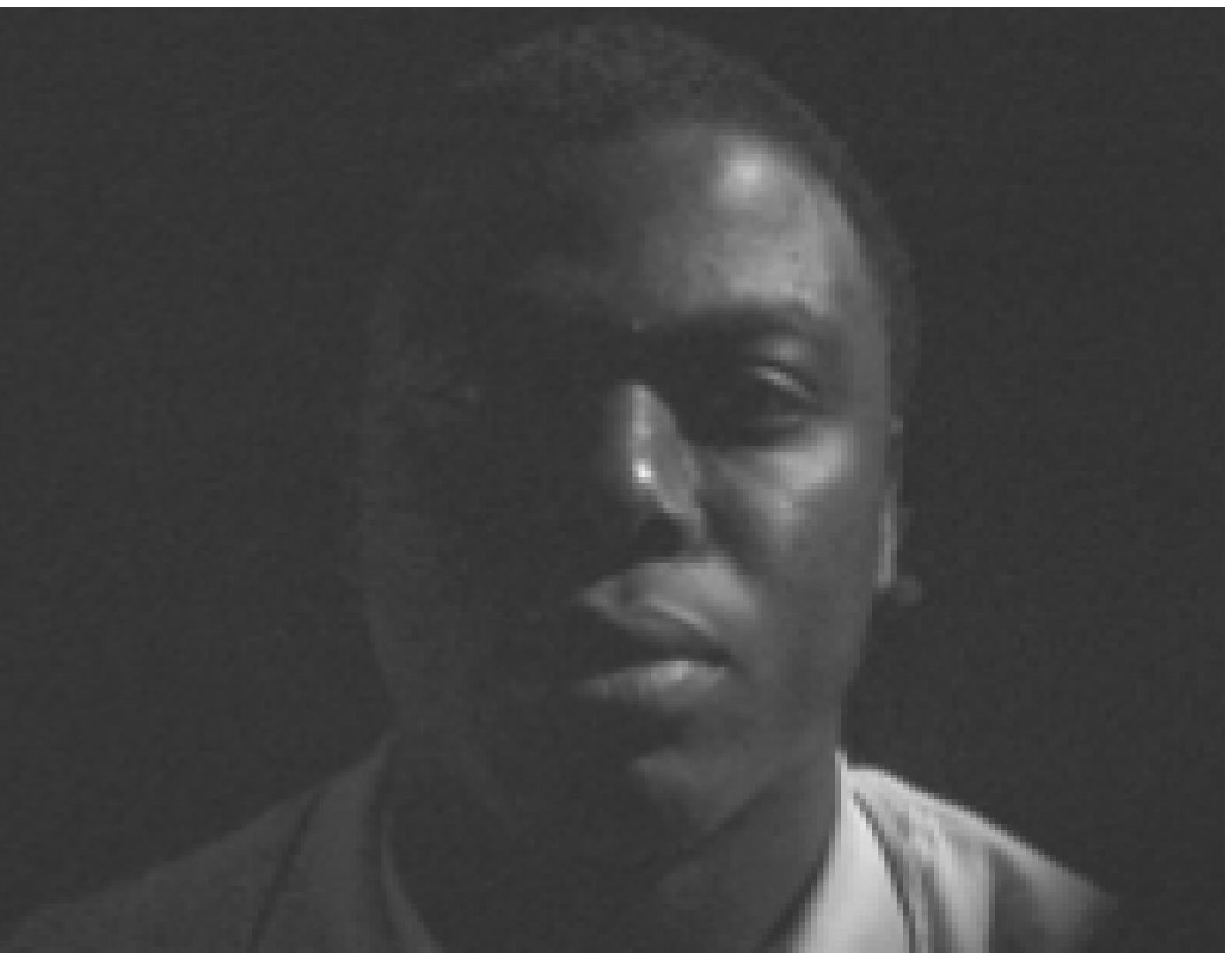}}
  \subfigure[540nm]{\includegraphics[width=0.22\textwidth]{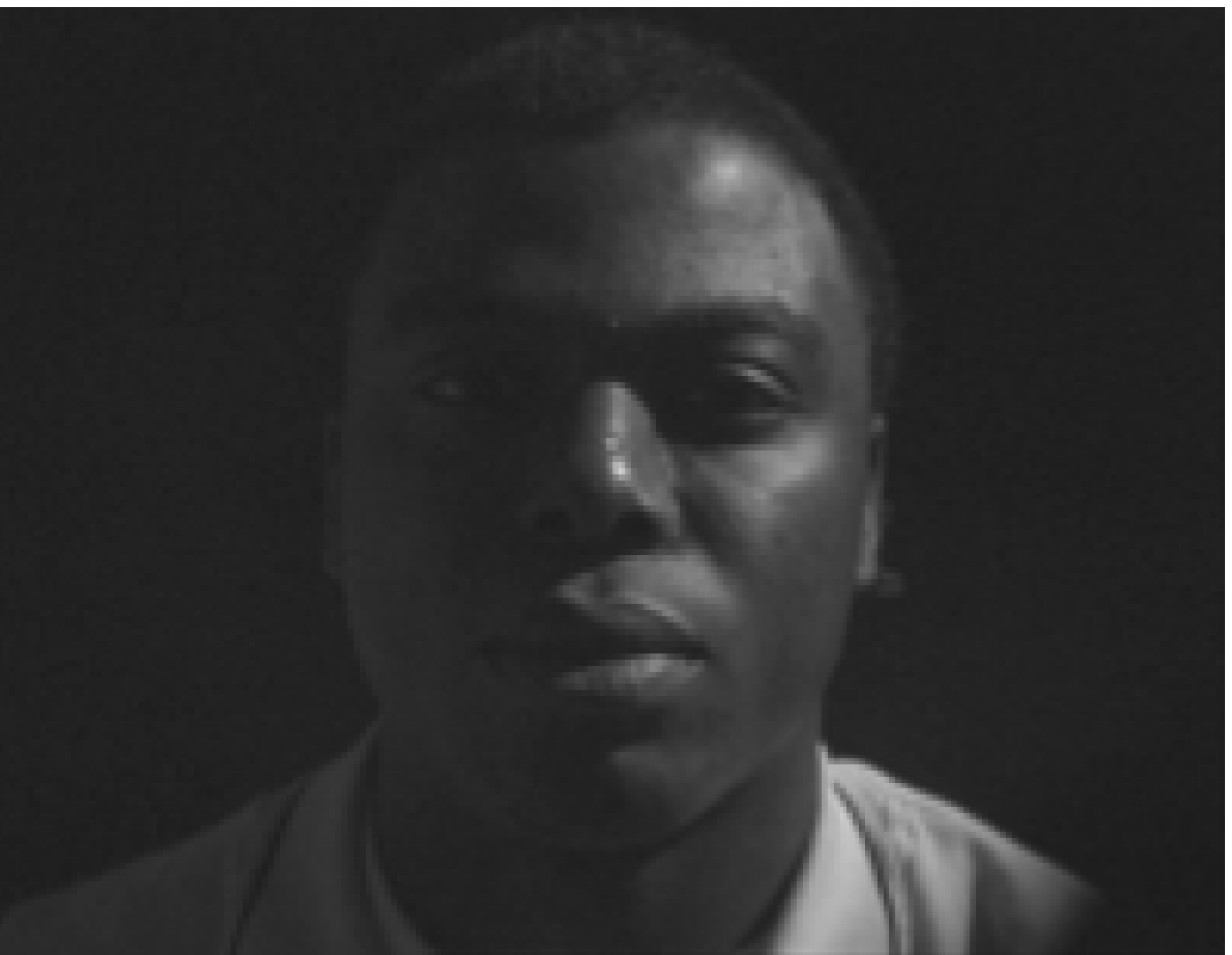}}
  \subfigure[600nm]{\includegraphics[width=0.22\textwidth]{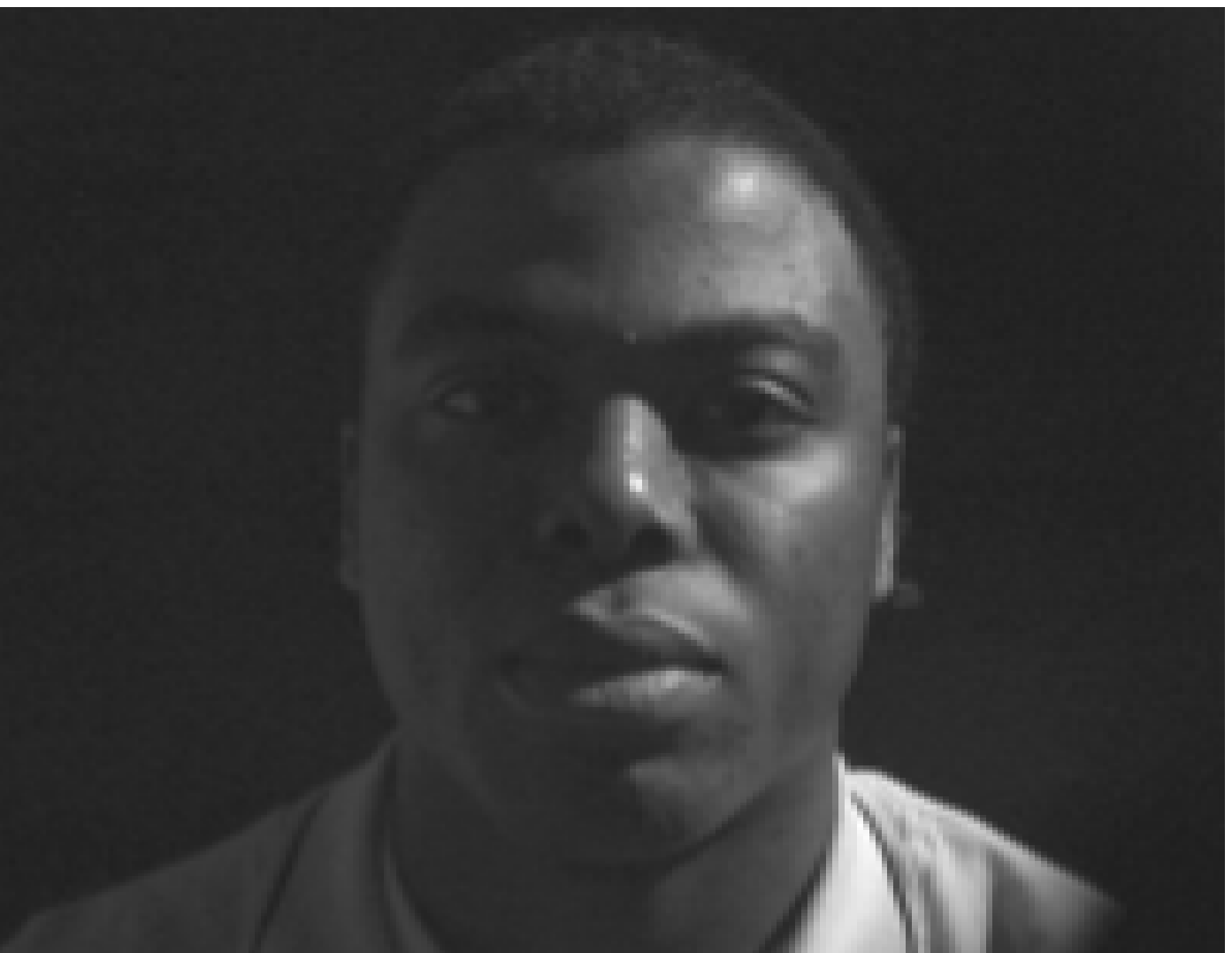}}
  \subfigure[660nm]{\includegraphics[width=0.22\textwidth]{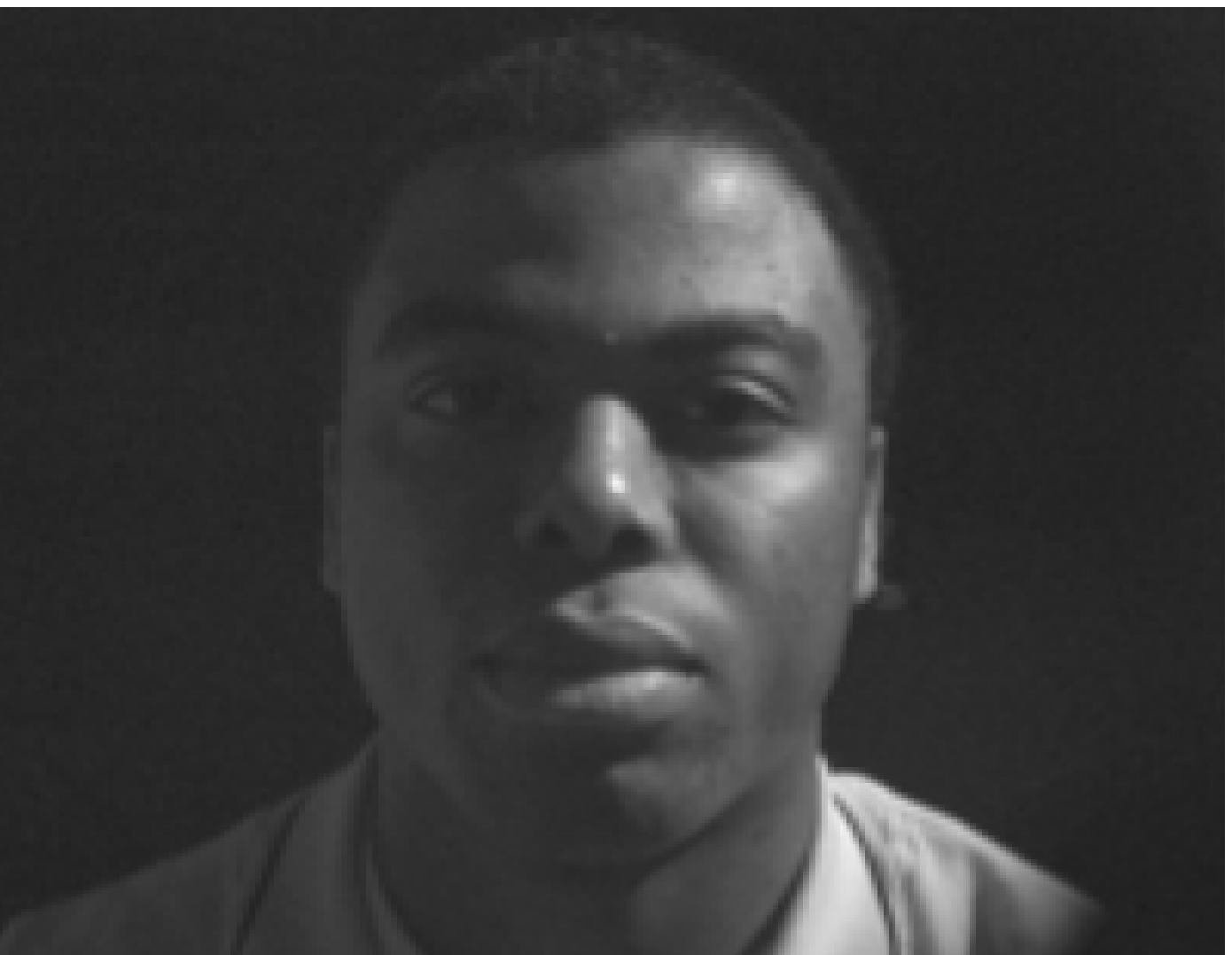}}
  \subfigure[720nm]{\includegraphics[width=0.22\textwidth]{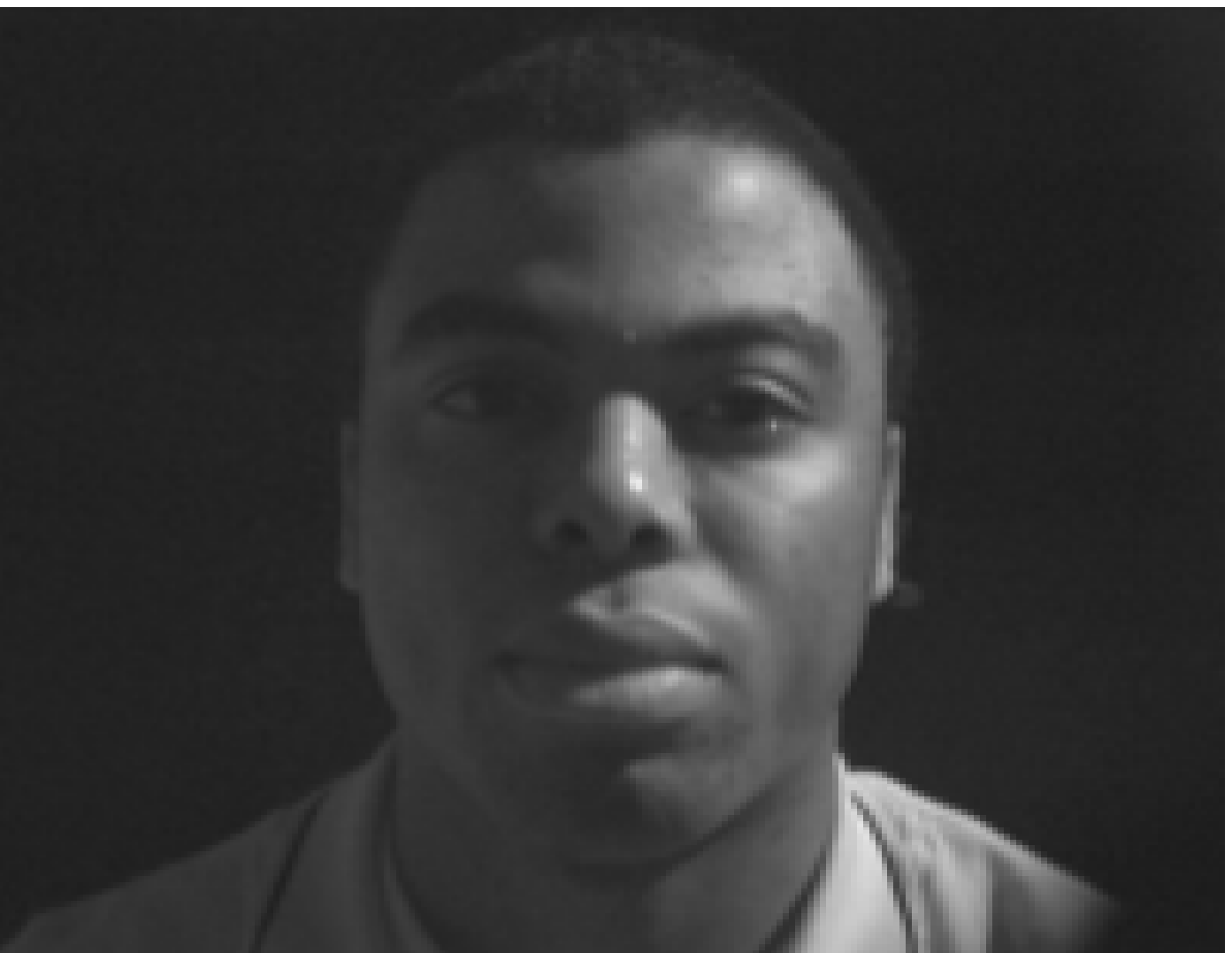}}
  \subfigure[Sunlight]{\includegraphics[width=0.22\textwidth]{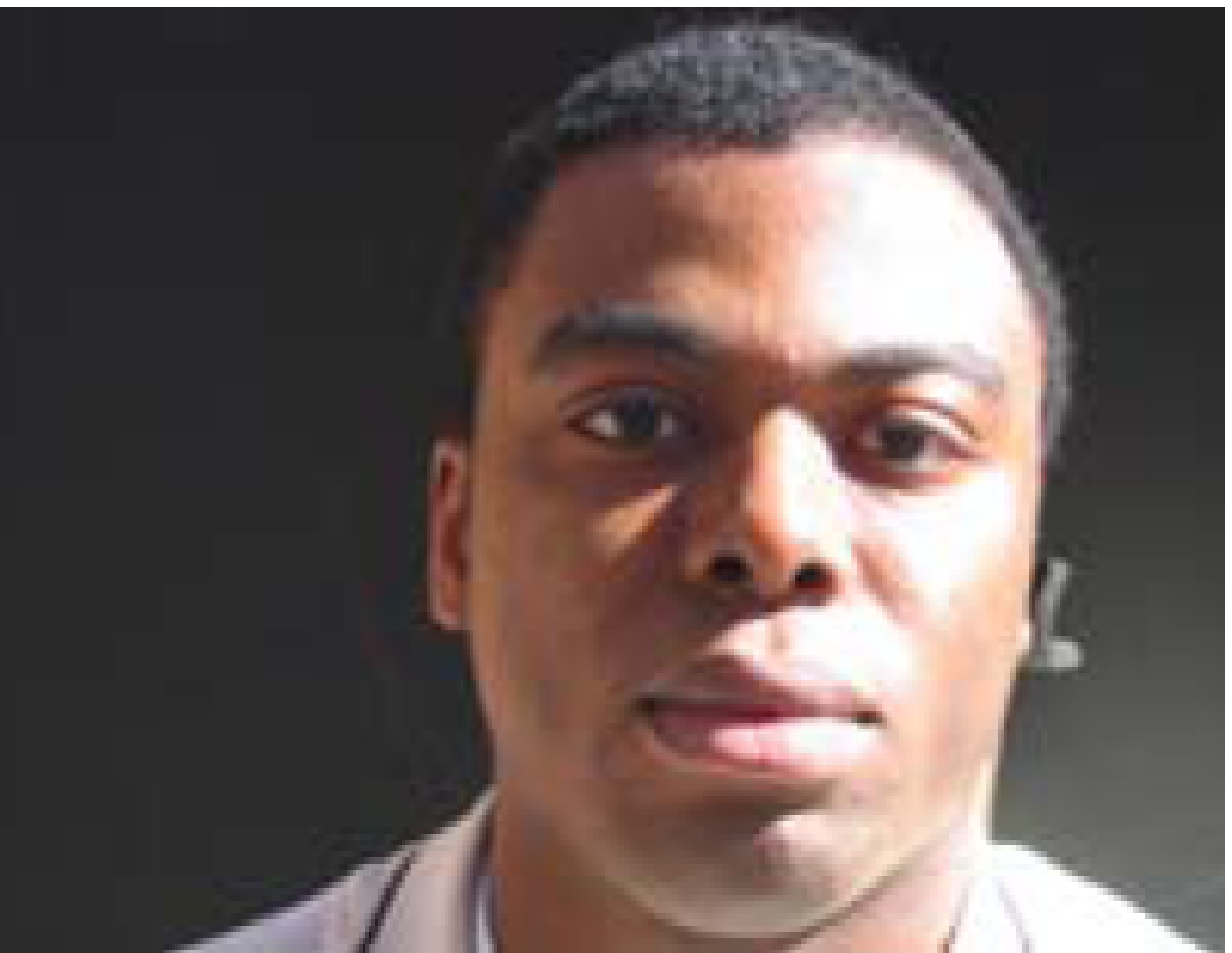}}
  \caption{ Eight images of a subject from the IRIS-M3 database. Shown are images acquired indoors in the (a) visible and (b) thermal spectrum, followed by (c,d,e,f,g)
            five multi-spectral images acquired in different sub-bands of the visible spectrum (these images are subtitled with the mean wavelength of the corresponding sub-band), and (f) an image acquired outdoors with natural daylight in a subsequent session. }
  \label{f:dbM3}
\end{figure}

The IRIS-M3 data set contains images of 82 people of various ethnicity, age and sex, and a total of 2624 images in $640 \times 480$ pixel resolution. Data was collected in two sessions. In the first session, which took place indoors, acquisition was performed under two illumination conditions: first using a halogen ambient lighting source and then a fluorescent ambient lighting source. Thus in both cases the faces were roughly homogeneously lit. In the second session, the acquisition of images was again performed under two illumination conditions: first using a fluorescent ambient lighting source indoors (as in the first session) and then outdoors in natural light. In the latter case, the subjects were oriented so that sunlight was illuminating their faces from a lateral direction. The IRIS-M3 data set does not contain any pose or expression variation: the subjects were asked to face the camera and maintain a neutral facial expression. Example images of a single subject from the database are shown in Fig.~\ref{f:dbM3}.

\subsection{University of Notre Dame (UND)}\label{ss:und}
\begin{figure}[htb]
  \centering
  \subfigure[Neutral expression]{\includegraphics[width=0.42\textwidth]{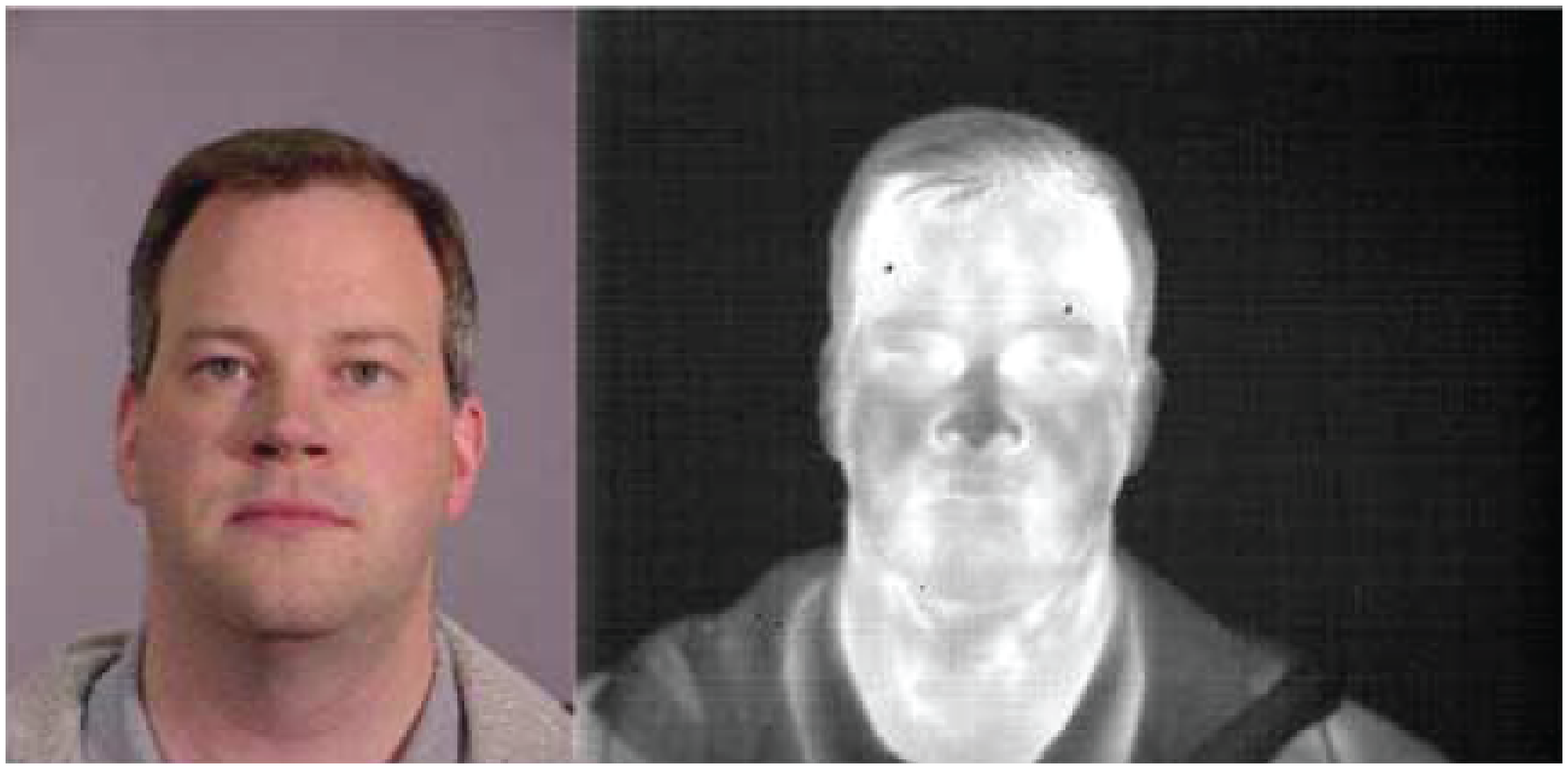}}\hspace{20pt}
  \subfigure[Smiling expression]{\includegraphics[width=0.42\textwidth]{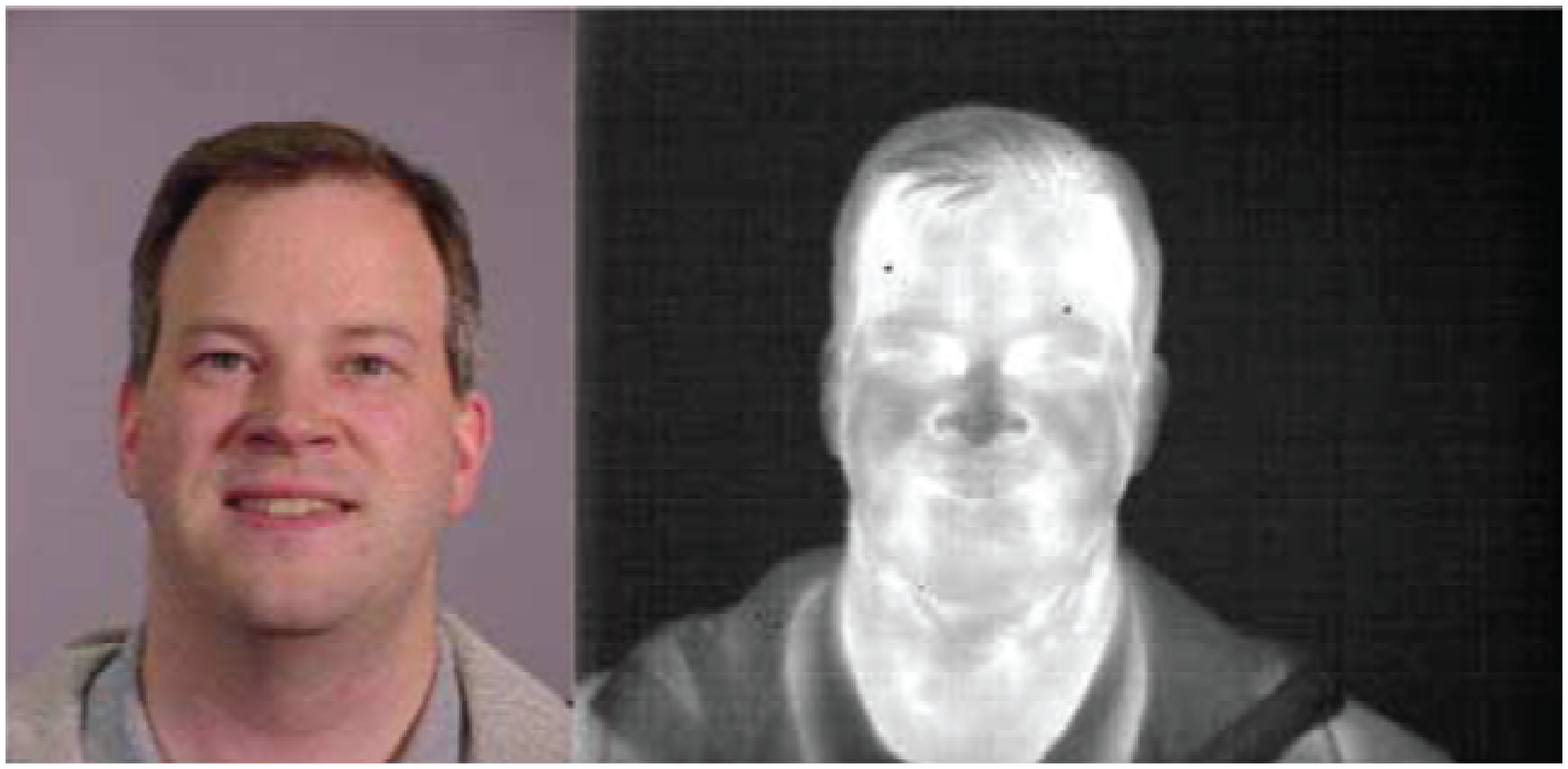}}
  \caption{ Visible and long wave infrared spectrum images of a person from the University of Notre Dame data set. The left-hand pair of images shows the subject in a neutral facial expression, while the right-hand pair shows the same subject smiling. }
  \label{f:dbUND}
\end{figure}

The University of Notre Dame data set (Collection C) \cite{DbUND} contains long wave infrared and visible spectrum images in $320 \times 240$ pixel resolution of 241 subjects under two illumination conditions. Three studio lights were used, one positioned in front of the subject and the other two in front and to the right and left of the subject. The first illumination in which data was acquired was obtained by having the frontal light off and the remaining lights off. The second illumination was obtained by having all lights switched on. For each illumination, two images were taken, one with the subject in a neutral facial expression and one smiling. Thus in each session four images per modality per subject were taken. Data was collected in multiple sessions in weekly intervals, different subjects participating in varying numbers of repeated sessions. The database contains a total of 2492 images some of which are shown in Fig.~\ref{f:dbUND}, and it is freely available upon request.

\subsection{University of Houston (UH)}\label{ss:uh}
\begin{figure}[htb]
  \centering
  \includegraphics[width=0.15\textwidth]{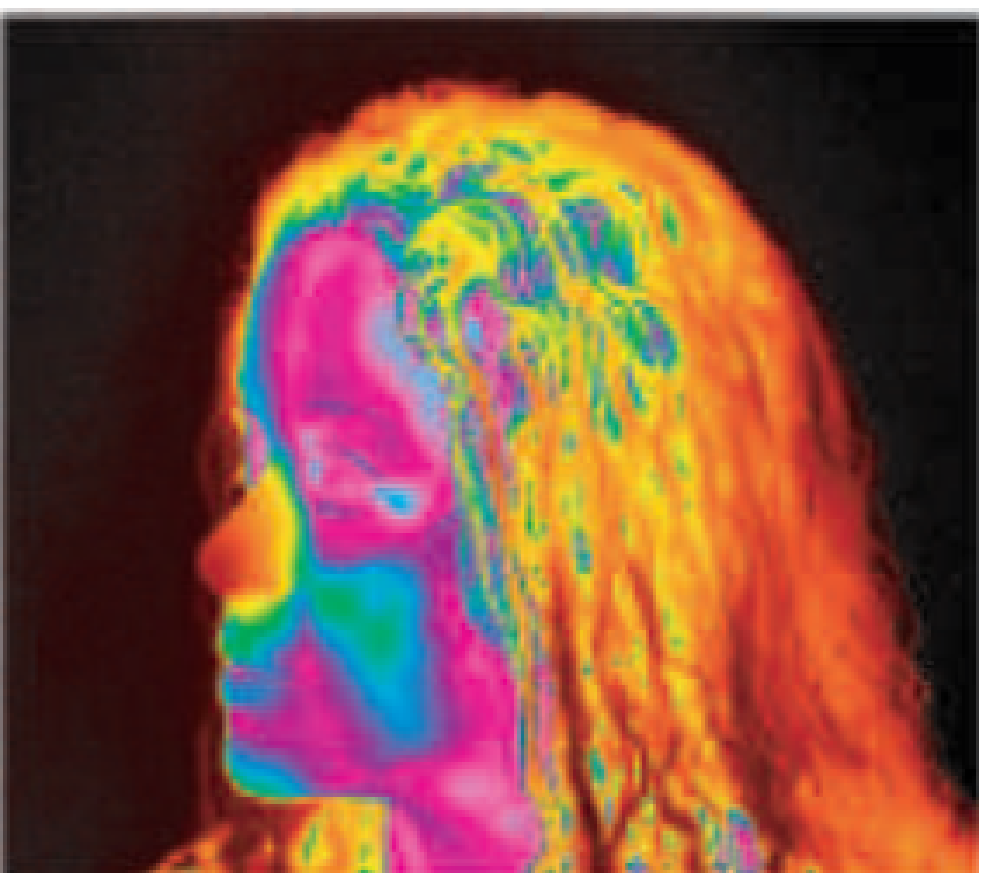}\hspace{20pt}
  \includegraphics[width=0.15\textwidth]{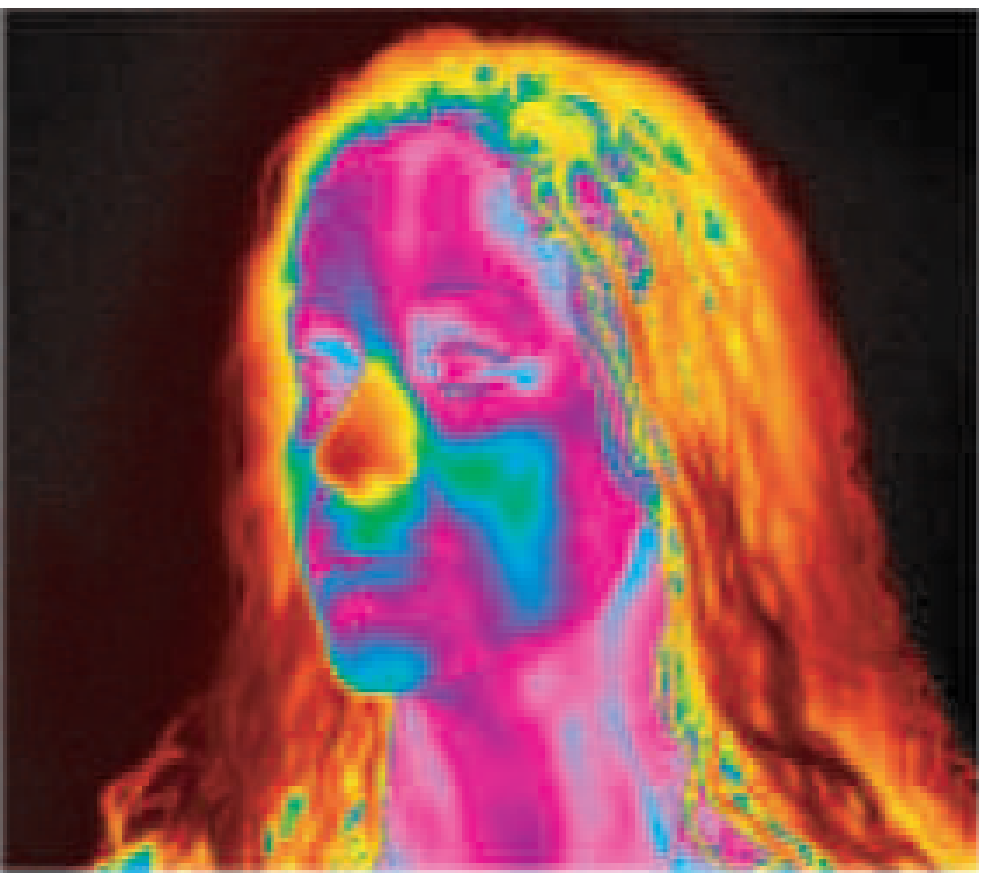}\hspace{20pt}
  \includegraphics[width=0.15\textwidth]{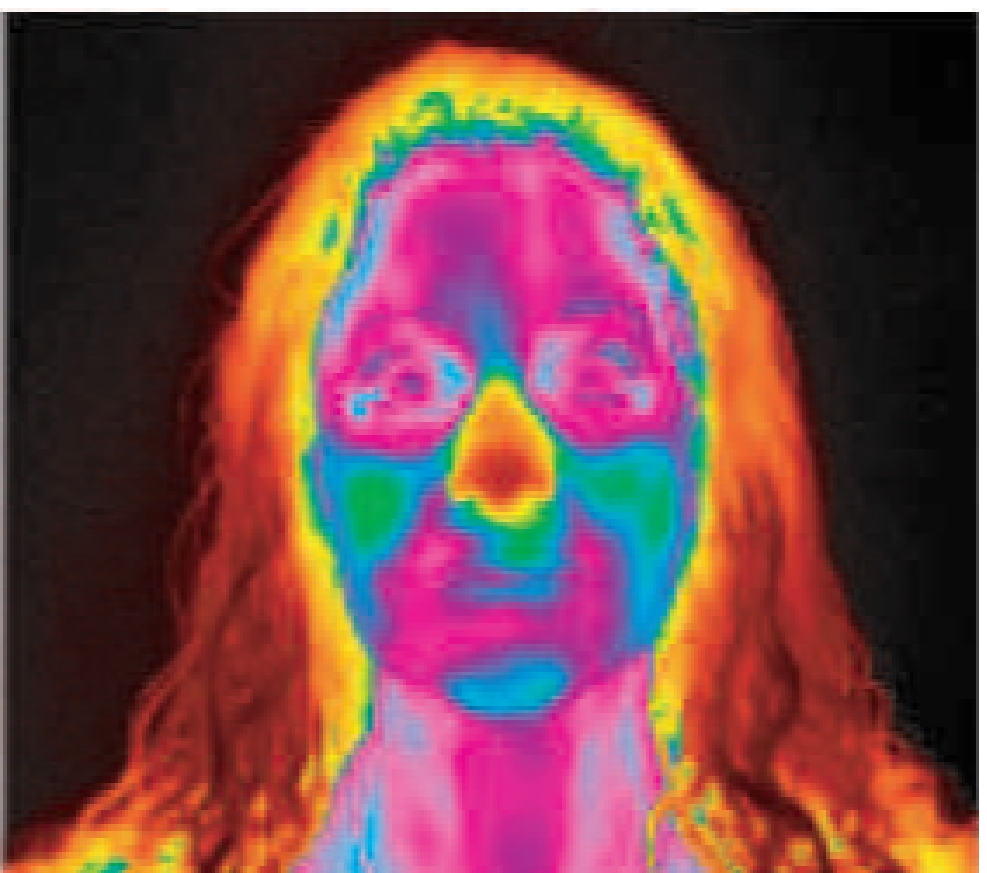}\hspace{20pt}
  \includegraphics[width=0.15\textwidth]{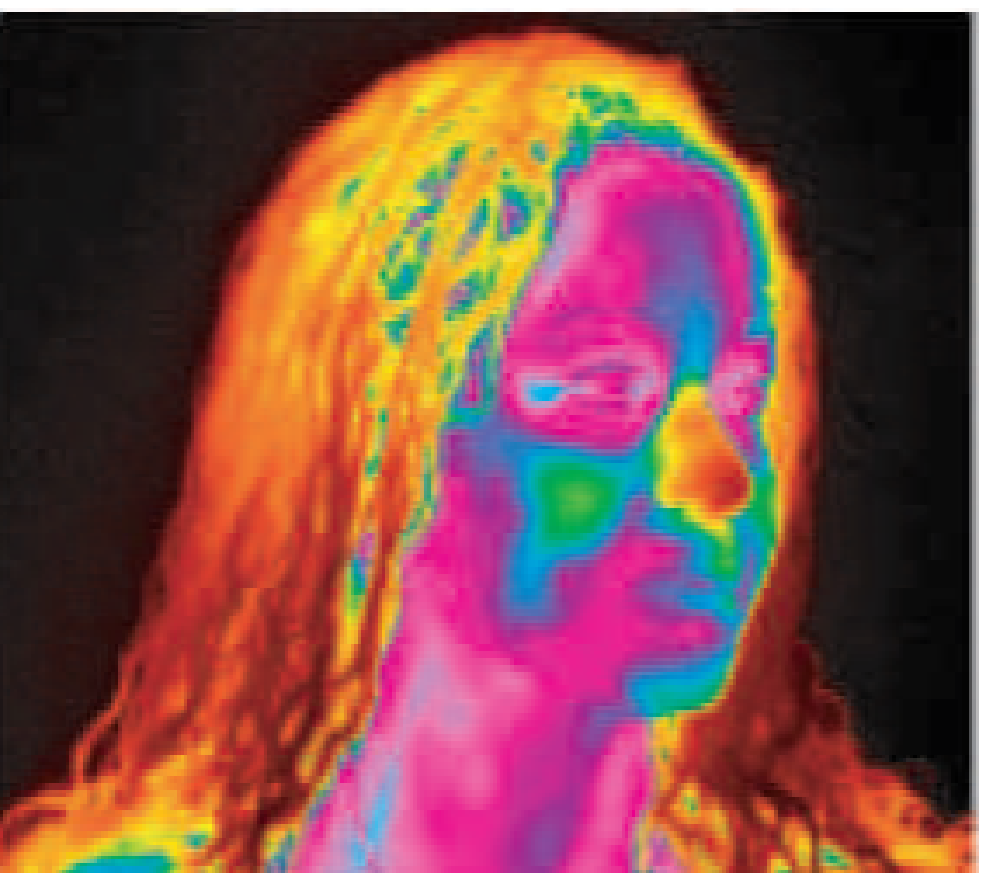}\hspace{20pt}
  \includegraphics[width=0.15\textwidth]{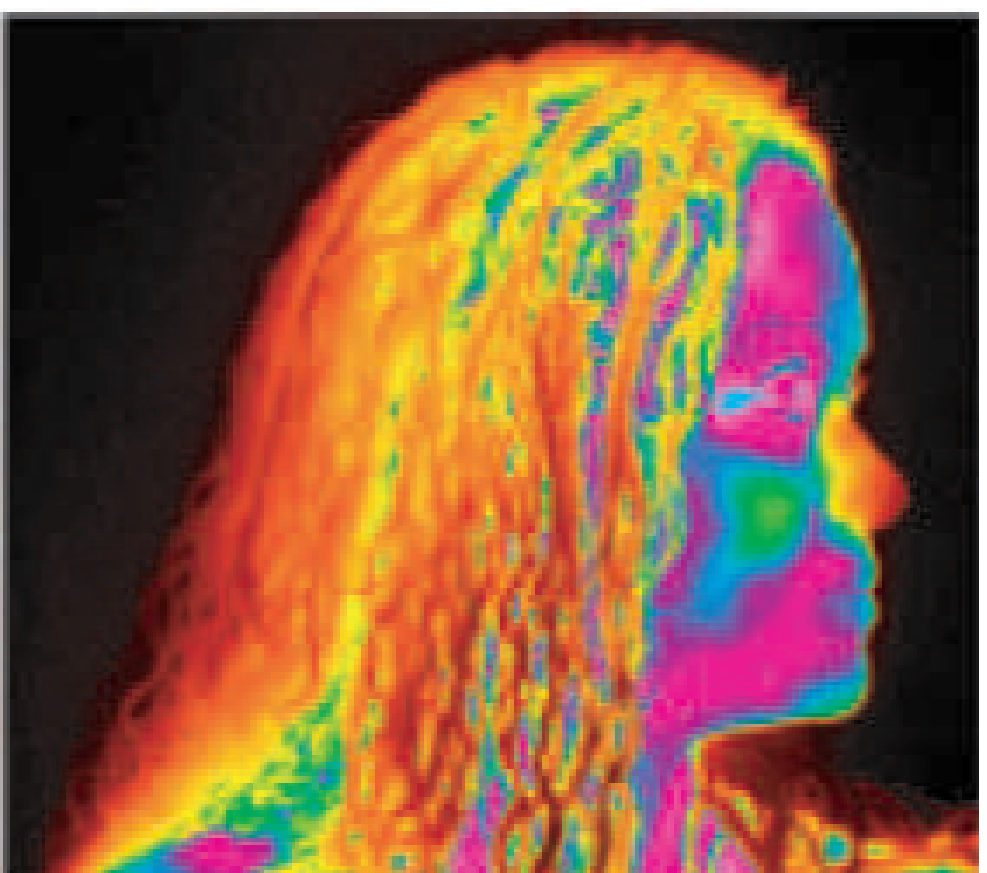}
  \caption{ False colour thermal appearance images of a subject in the five key poses in the University of Houston data set. }
  \label{f:dbUH}
\end{figure}

The University of Houston database consists of a total of 7590 thermal images of 138 subjects, with a uniform distribution of 55 images per subject. Subjects are of various ethnicity, age and sex. With the exception of four subjects, from whom data was collected in two sessions six months apart, the data for a particular subject was acquired in a single session. The exact protocol which was used to introduce pose and expression variability in the data set was not described by the authors \cite{BuddPavlTsiaBaza2007}. Example images are shown in Fig.~\ref{f:dbUH}. The database is available free of charge upon request.

\subsection{Surveillance Cameras Face Database (SCface)}\label{ss:scface}
\begin{figure}[htb]
  \centering
  \includegraphics[height=0.18\textwidth]{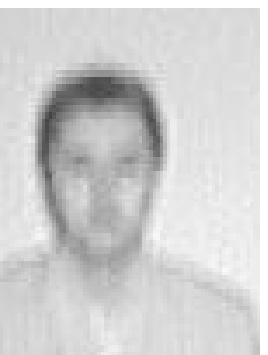}\hspace{10pt}
  \includegraphics[height=0.18\textwidth]{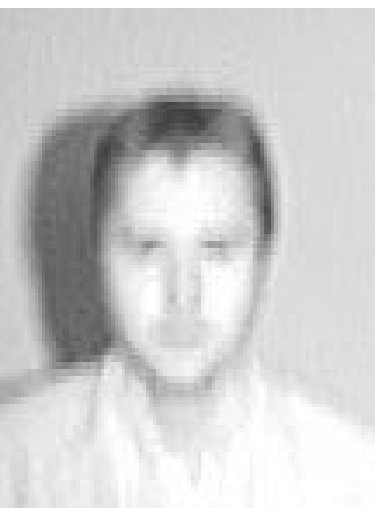}\hspace{10pt}
  \includegraphics[height=0.18\textwidth]{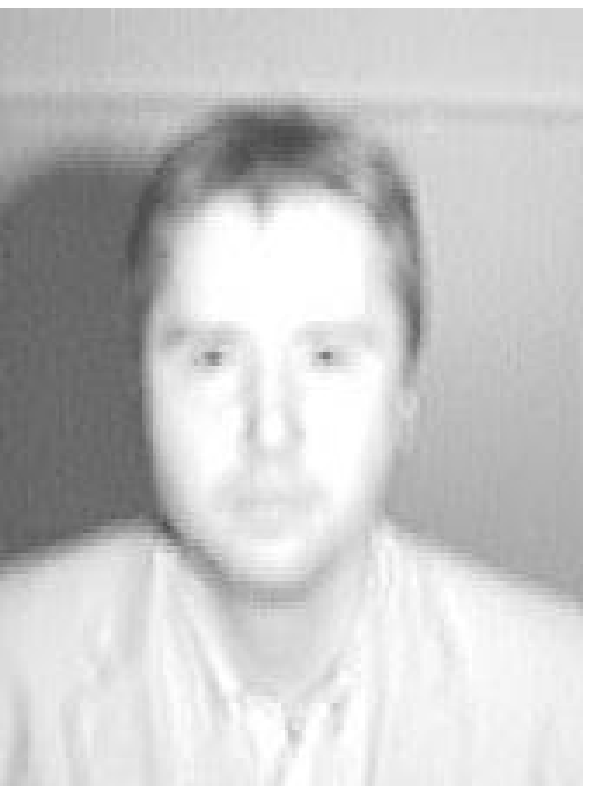}\hspace{10pt}
  \includegraphics[height=0.18\textwidth]{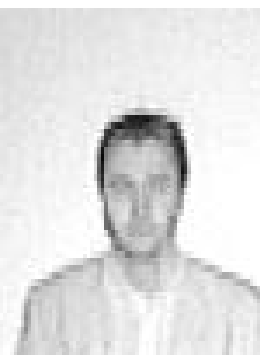}\hspace{10pt}
  \includegraphics[height=0.18\textwidth]{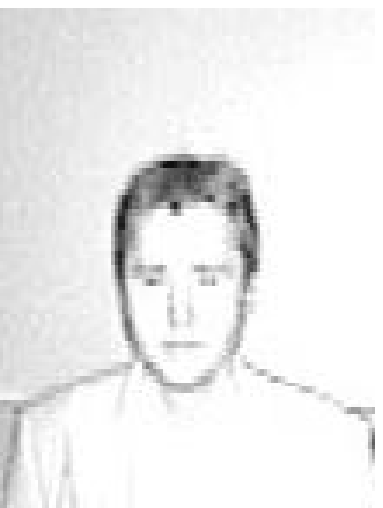}\hspace{10pt}
  \includegraphics[height=0.18\textwidth]{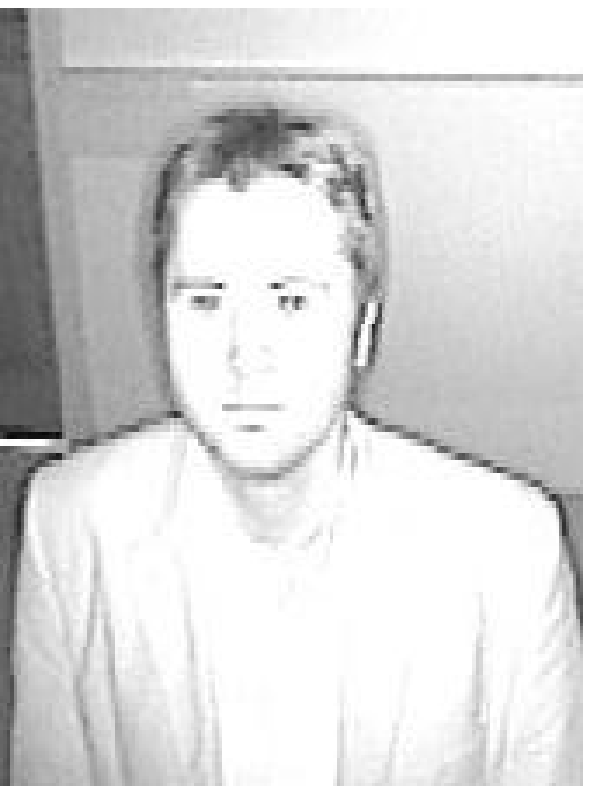}
  \caption{ A set of images from the Surveillance Cameras Face Database \cite{GrgiDelaGrgi2011} collected at the University of Zagreb.  Images were collected in a ``real-world'' setup using a set of surveillance cameras of different resolutions and quality. Illumination, pose and facial expression of subjects (University staff) were not explicitly controlled. }
  \label{f:dbSCface}
\end{figure}

The Surveillance Cameras Face Database \cite{GrgiDelaGrgi2011} is a particulary interesting data set because it was acquired using a setup substantially different from those adopted for the collection of other infrared databases described here. SCface has only recently been made public which is why it was not used in any of the publications reviewed herein. Of all the publicly available databases, the variability of extrinsic factors such as illumination or pose in this data set is controlled the least. Images were collected in a ``real-world'' setup using a set of visual and thermal spectrum surveillance cameras imaging hallways of a University of Zagreb building. Thus illumination, pose, camera resolution, face scale (distance from the camera) and to a lesser degree facial expression are all variable. The data set contains 130 subjects and the total of 4160 images collected over five days see Fig.~\ref{f:dbSCface}.

\subsection{Florida State University (FSU)}\label{ss:fsu}
\begin{figure}[htb]
  \centering
  \includegraphics[width=0.95\textwidth]{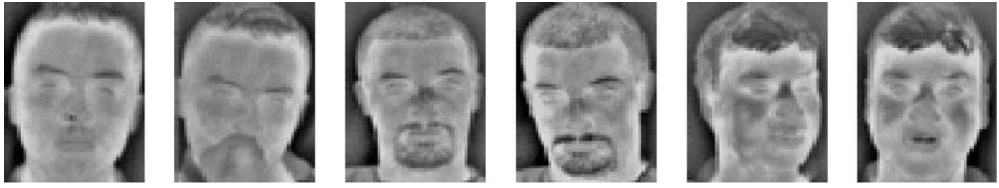}
  \caption{ Examples of images from the Florida State University infrared database showing typical pose and facial expression variability in the data set. }
  \label{f:dbFSU}
\end{figure}

The publicly available face data set collected at Florida State University comprises 234 images in $320 \times 240$ pixel resolution of 10 different subjects across a range of \textit{ad lib} adopted poses and facial expressions \cite{SrivLiuThomHesh2001}, as illustrated in Fig.~\ref{f:dbFSU}. It is available for download at \url{http://lcv.stat.fsu.edu}.

\subsection{UC Irvine Hyperspectral (UC)}\label{ss:uc}
\begin{figure}[htb]
  \centering
  \subfigure[Pose variation]{\includegraphics[height=0.15\textwidth]{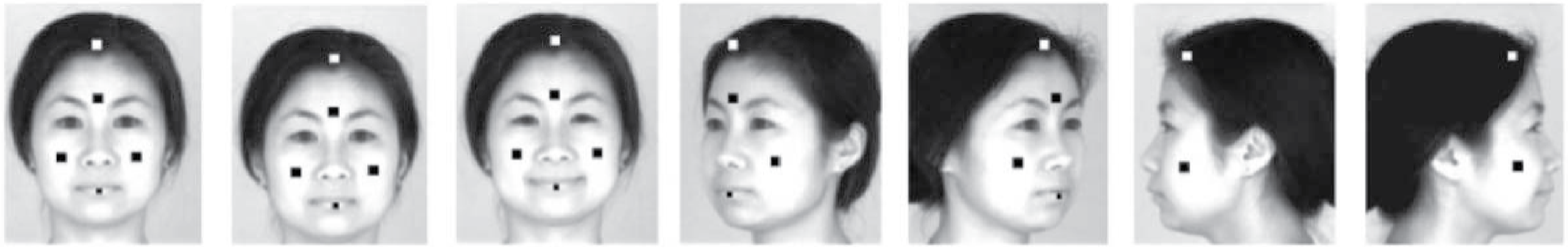}}
  \subfigure[Person 1: seven multi-spectral images]{\includegraphics[height=0.15\textwidth]{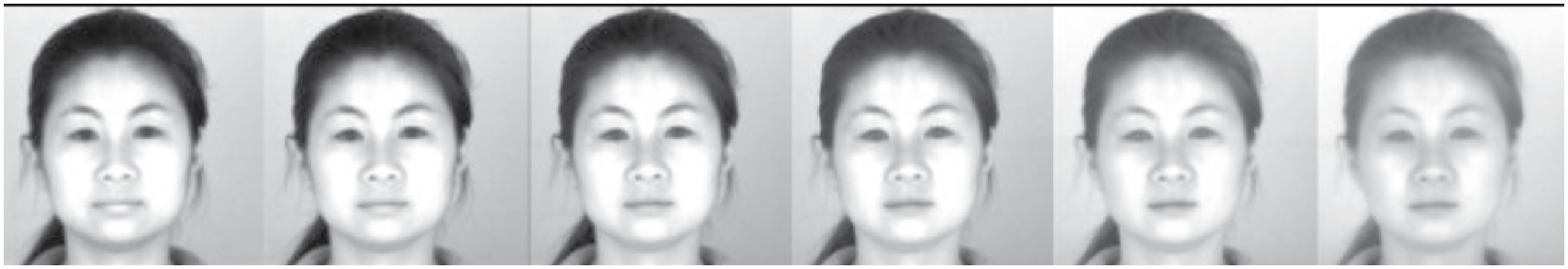}}
  \subfigure[Person 2: seven multi-spectral images]{\includegraphics[height=0.15\textwidth]{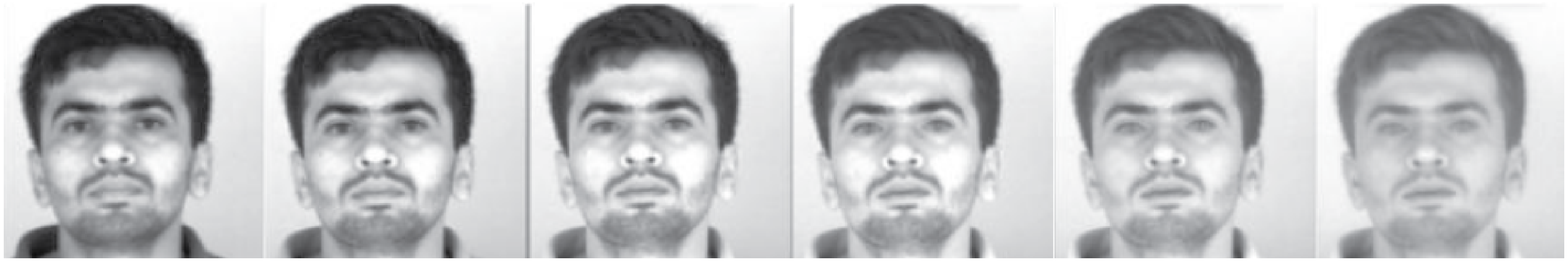}}
  \caption{ (a) For the UC Irvine Hyperspectral data set subjects were imaged with a neutral facial expression in five different poses (the frontal pose twice) and
            smiling in the frontal pose. For each pose/expression combination, multi-spectral images were acquired in 0.1$\mu$m wide sub-bands of the near infrared spectrum. (b) Multi-spectral images corresponding to seven equidistant (wavelength-wise) sub-bands spanning the near infrared spectrum are shown here. }
  \label{f:dbUC}
\end{figure}

The University of California/Irvine collected a data set of multi-spectral images of 200 subjects. All images were captured in $468 \times 494$ pixel resolution \cite{PanHealPrasTrom2003} and under halogen ambient illumination. Subjects were imaged with a neutral facial expression in the frontal, and two profile and semi-profile poses, as well as with a smiling expression in the frontal pose only. For each pose and illumination 31 multi-spectral images were captured for 0.1$\mu$m wide sub-bands of the near infrared spectrum. For twenty of the 200 subjects, data acquisition was repeated after a time lapse of up to five weeks. Fig.~\ref{f:dbUC}~(a) shows images of a subject for different poses and facial expressions, with multi-spectral images of two subjects in seven equidistant sub-bands covering the near wave infrared spectrum in Fig.~\ref{f:dbUC}~(b,c).

\subsection{West Virginia University Multispectral (WVUM)}\label{ss:WVUM}
\begin{figure}[htb]
  \centering
  \subfigure[950nm]{\includegraphics[width=0.18\textwidth]{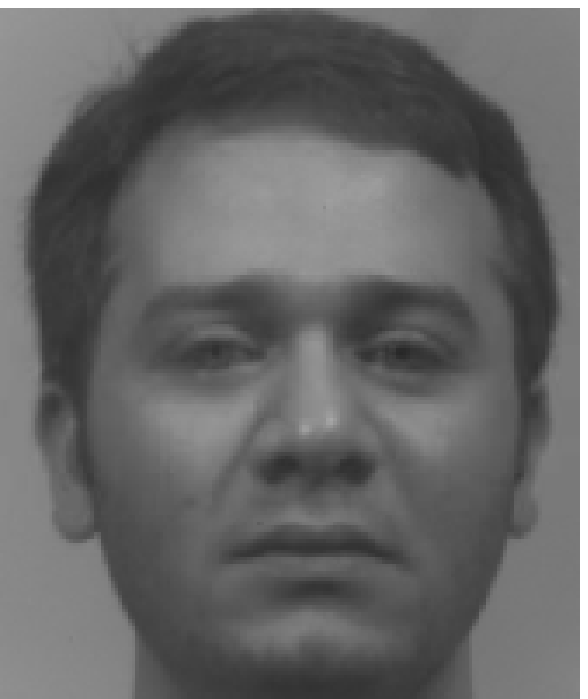}}\hspace{5pt}
  \subfigure[1150nm]{\includegraphics[width=0.18\textwidth]{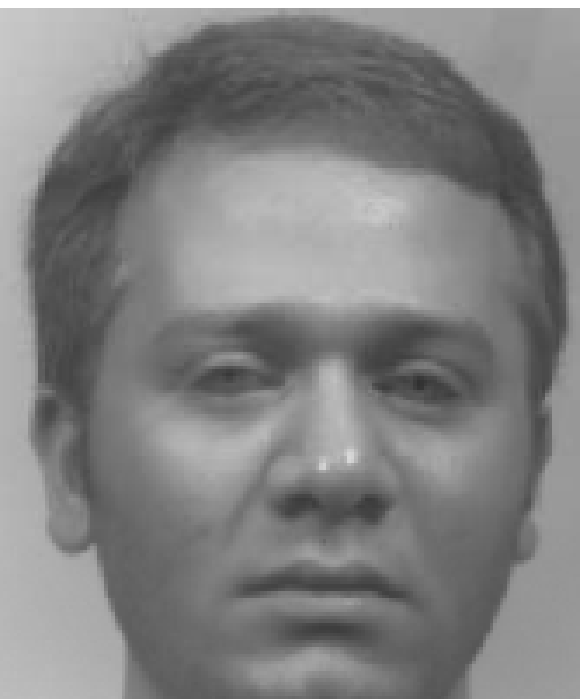}}\hspace{5pt}
  \subfigure[1350nm]{\includegraphics[width=0.18\textwidth]{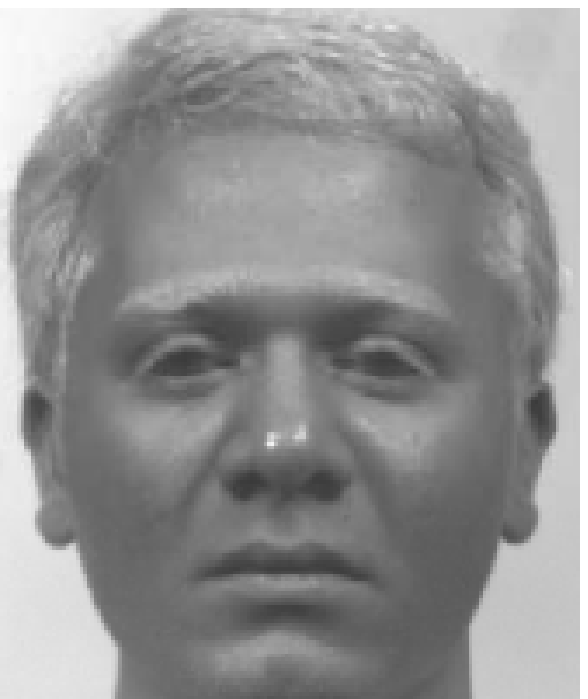}}\hspace{5pt}
  \subfigure[1550nm]{\includegraphics[width=0.18\textwidth]{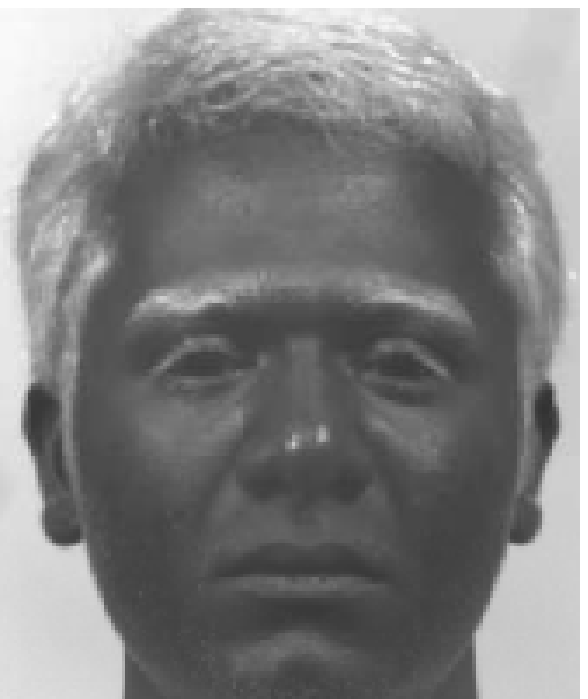}}\hspace{5pt}
  \subfigure[Visible]{\includegraphics[width=0.18\textwidth]{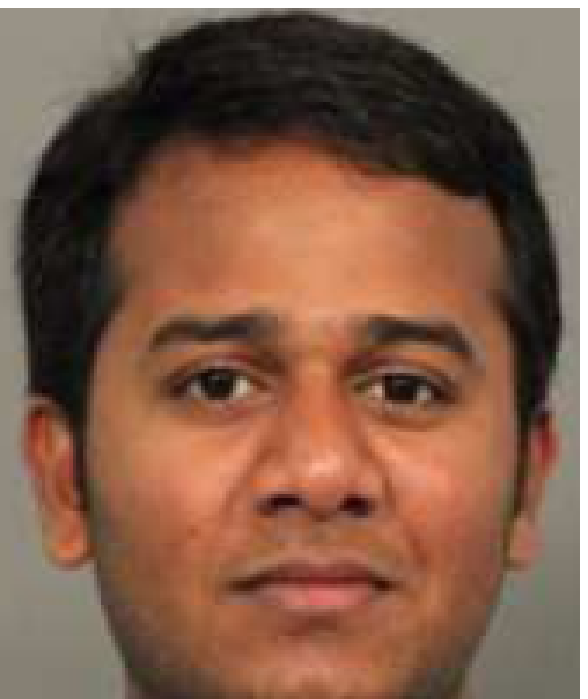}}
  \caption{ Examples of images from the West Virginia University Multispectral database. Shown are matching images acquired in different spectral sub-bands. }
  \label{f:dbWVUM}
\end{figure}

The West Virginia University Multi-spectral database consists of visible and short wave infrared spectrum images of 50 subjects. In the visible spectrum, 25 frontal face images were captured for each subject in the database, giving the total of 1250 images. In the short wave infrared spectrum, faces were imaged in the frontal, and left and right semi-profile (67.5$^{\circ}$ from frontal) poses. For each pose nine multi-spectral images were acquired corresponding to 100nm wide spectral sub-bands in the range from 950nm to 1650nm. Thus there are 1350 short wave infrared images in the database. Data for each person was collected in two sessions, up to a month apart. Example images are shown in Fig.~\ref{f:dbWVUM}.

\subsection{The Hong Kong Polytechnic University NIR Face Database (PolyU-NIR)}\label{ss:PolyU}
\begin{figure}[htb]
  \centering
  \subfigure[]{\includegraphics[width=0.23\textwidth]{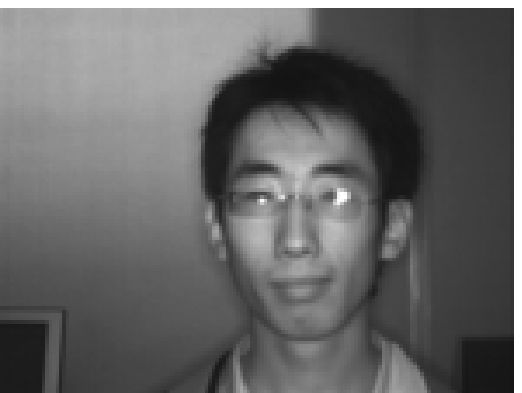}}\hspace{5pt}
  \subfigure[]{\includegraphics[width=0.23\textwidth]{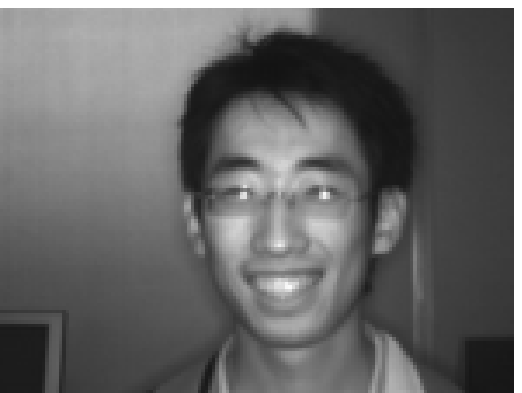}}\hspace{5pt}
  \subfigure[]{\includegraphics[width=0.23\textwidth]{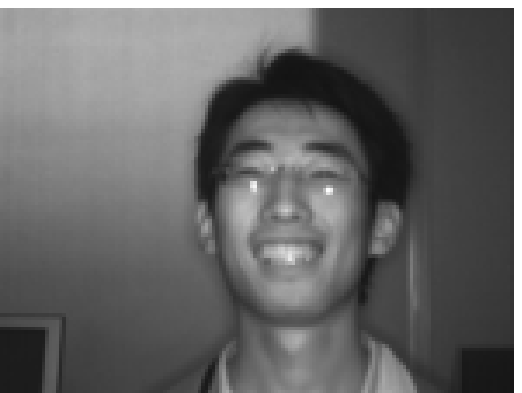}}\hspace{5pt}
  \subfigure[]{\includegraphics[width=0.23\textwidth]{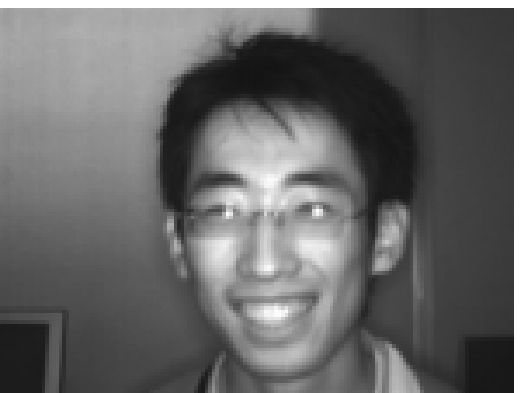}}
  \caption{ Examples of images from the Hong Kong Polytechnic University NIR Face Database. Shown are images of a single subject across a moderate degree of scale, pose and facial expression variation. }
  \label{f:dbPolyU}
\end{figure}

The Hong Kong Polytechnic University NIR Face Database is one of the few freely available data sets which contains images of faces acquired in the NIR sub-band of the infrared spectrum. It contains approximately 34,000 images of 335 individuals with a moderate degree of scale, pose and facial expression variation within the data subset of each subject. Example images are shown in Fig.~\ref{f:dbPolyU}. More information on the database can be obtained from the original publication \cite{ZhanZhanZhanShen2010} and at \url{http://www4.comp.polyu.edu.hk/~biometrics/polyudb_face.htm}.

\subsection{The Laval University Thermal IR Face Motion Database }\label{ss:dbULaval}
\begin{figure*}[htb]
  \centering
  \includegraphics[width=1\textwidth]{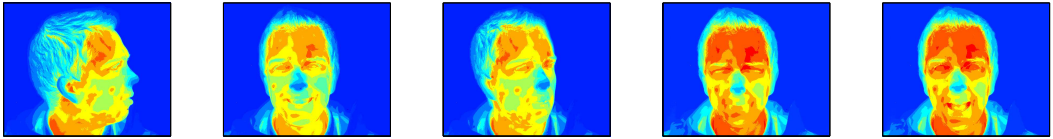}
  \caption{ False colour thermal appearance images of a subject in five arbitrary poses and facial expressions in the Laval University Thermal IR Face Motion data set. }
  \label{f:dbFM}
\end{figure*}

The Laval University Thermal IR Face Motion Database is the only freely available data set which contains videos of faces acquired in the IR spectrum. It contains 200 individuals of varying age, ethnicity and gender, with two sequences collected for each person. Each video sequence is 10~s long and was captured at 30~fps, thus resulting in 300 frames of 640 $\times$ 512 pixels. The imaged subjects were instructed to perform head motion that covers the yaw range from frontal (0 degrees) approximately full profile (90 degrees) face orientation relative to the camera, without any special attention to the tempo of the motion or the time spent in each pose. The subjects were also asked to display an arbitrary range of facial expressions. Examples of frames from a single video sequence are shown in Fig.~\ref{f:dbFM}. The data set if freely available for research purposes and can be obtained by contacting the authors \cite{GhiaAranBendMald2013a}.

\section{Summary and Conclusions}\label{s:summary}
Systems based on images acquired in the visible spectrum have reached a significant level of maturity with some yet limited practical success. A range of nuisance factors continue to pose serious problems when visible spectrum based face recognition methods are applied in a real-world setting. Dealing with illumination, pose and facial expression changes, and facial disguises is still a major challenge. The use of infrared imaging which has emerged as an alternative to visual spectrum based approaches, has attracted substantial research and commercial attention as a modality which could facilitate greater robustness to illumination and facial expression changes, facial disguises and dark environments. On the other hand, both theoretical and empirical evidence reveals a number of nuisance variables which affect infrared appearance too. These include occlusion by corrective eyeglasses, the person's emotional state, postprandial thermogenesis, alcohol consumption and compensatory bodily temperature changes to ambient temperature. Early work on infrared based face recognition which mostly explored the use of standard statistical techniques applied on holistic appearance was generally unsuccessful in dealing with the aforementioned challenges when applied on realistic data. Feature based methods, which extract more robust facial biometric characteristics of a face from infrared images have been more successful. Particularly interesting are methods which are based on the distribution of superficial vessels. Vascular network based method extract and use this information explicitly, while blood perfusion based methods synthesize quasi-invariant images using peripheral blood flow models. We also expect that algorithms based on the concept of sparse representation, which have recently received a significant amount of attention in the sphere of visible spectrum-based face recognition \cite{WrigYangGaneSast+2009,DengHuGuo2012}, could offer interesting insights when applied on IR data.

A notable limitation which we found in all of the reviewed publications, is of a methodological nature: despite the universal acknowledgment of the major challenges of infrared based face recognition, none of the reported experiments evaluate performance in the context of all of them. What is more, even papers which do use a public database often perform evaluation on only a subset of the data. These reasons make a direct comparison of different approaches difficult, as well as the assessment of their capacity for practical deployment. Consequently we encourage efforts directed towards the collection of large scale data sets acquired in realistic conditions and their consistent use in the evaluation and reporting of novel recognition algorithms. Indeed, in this paper we also reviewed a range of data sets currently available to researchers.

Considering the results published to date, in the opinion of these authors two particularly promising ideas stand out: (i) the development of identity descriptors based on persistent physiological features such as vascular networks, and (ii) the use of methods for multi-modal fusion of complementary data types, most notably those based on visible and infrared images. Both are still in their early stages, with a potential for significant further improvement.

\begin{spacing}{1.0}
  \bibliographystyle{unsrt}
  \bibliography{./2013_PR_paper2_NEW,./my_bibliography}
\end{spacing}

\end{document}